\documentclass[10pt,twocolumn,letterpaper,table]{article}

\usepackage{cvpr}
\usepackage{times}
\usepackage{epsfig}
\usepackage{graphicx}
\usepackage{amsmath}
\usepackage{amssymb}

\usepackage[utf8]{inputenc}
\usepackage{enumitem}
\usepackage{xcolor}
\usepackage[skip=4pt]{subcaption}
\usepackage{adjustbox}
\usepackage{multirow}
\usepackage{hhline}

\usepackage[pagebackref=true,breaklinks=true,letterpaper=true,colorlinks,bookmarks=false]{hyperref}

\usepackage[sort&compress,capitalize,noabbrev]{cleveref}
\creflabelformat{equation}{#2#1#3}

\makeatletter
\renewcommand{\paragraph}{%
  \@startsection{paragraph}{4}%
  {\z@}{1.75ex \@plus 1ex \@minus .2ex}{-1em}%
  {\normalfont\normalsize\bfseries}%
}
\makeatother

\newcommand\Tstrut{\rule{0pt}{2.6ex}}         % = `top' strut
\newcommand\Bstrut{\rule[-0.8ex]{0pt}{0pt}}   % = `bottom' strut

\cvprfinalcopy % *** Uncomment this line for the final submission

\def\cvprPaperID{576} % *** Enter the CVPR Paper ID here

\ifcvprfinal\fi
\begin{document}

%%%%%%%%% TITLE
\title{SDC -- Stacked Dilated Convolution:\\A Unified Descriptor Network for Dense Matching Tasks}

\author{René Schuster\textsuperscript{1} \hspace{0.5cm} Oliver Wasenmüller\textsuperscript{1} \hspace{0.5cm} Christian Unger\textsuperscript{2} \hspace{0.5cm} Didier Stricker\textsuperscript{1}\\
\textsuperscript{1}DFKI - German Research Center for Artificial Intelligence \hspace{0.5cm} \textsuperscript{2}BMW Group \\
{\tt\small firstname.lastname@\string{bmw,dfki\string}.de}
% \and
% Christian Unger\\
% BMW Group\\
% {\tt\small christian.unger@bmw.de}
}

\maketitle
\ifcvprfinal\thispagestyle{empty}\fi

%%%%%%%%% ABSTRACT
\begin{abstract}
Dense pixel matching is important for many computer vision tasks such as disparity and flow estimation. We present a robust, unified  descriptor network that considers a large context region with high spatial variance. Our network has a very large receptive field and avoids striding layers to maintain spatial resolution. These properties are achieved by creating a novel neural network layer that consists of multiple, parallel, stacked dilated convolutions (SDC). Several of these layers are combined to form our SDC descriptor network. In our experiments, we show that our SDC features outperform state-of-the-art feature descriptors in terms of accuracy and robustness. In addition, we demonstrate the superior performance of SDC in state-of-the-art stereo matching, optical flow and scene flow algorithms on several famous public benchmarks.
\end{abstract}

%%%%%%%%% BODY TEXT
\section{Introduction} \label{sec:intro}
\begin{figure}[t]
	\begin{center}
		\begin{subfigure}[c]{0.04\columnwidth}
			\rotatebox[origin=c]{90}{\small ELAS \cite{geiger2010elas}}%
		\end{subfigure}
		\begin{subfigure}[c]{0.47\columnwidth}
			\includegraphics[width=\textwidth]{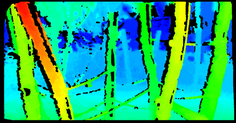}
		\end{subfigure}
		\begin{subfigure}[c]{0.47\columnwidth}
			\includegraphics[width=\textwidth]{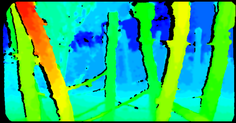}
		\end{subfigure}\\%
		\vspace{1mm}%
		\begin{subfigure}[c]{0.04\columnwidth}
			\rotatebox[origin=c]{90}{\small CPM \cite{hu2016efficient}}%
		\end{subfigure}
		\begin{subfigure}[c]{0.47\columnwidth}
			\includegraphics[width=\textwidth]{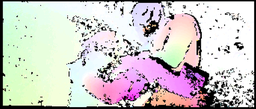}
		\end{subfigure}
		\begin{subfigure}[c]{0.47\columnwidth}
			\includegraphics[width=\textwidth]{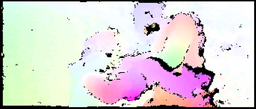}
		\end{subfigure}\\%
		\vspace{1mm}%
		\begin{subfigure}[t]{0.04\columnwidth}
			\rotatebox[origin=c]{90}{\small SFF \cite{schuster2018sceneflowfields}}%
		\end{subfigure}
		\begin{subfigure}[t]{0.47\columnwidth}
			\adjincludegraphics[width=\textwidth,trim={{0.1\width} 0 {0.2\width} 0},clip]{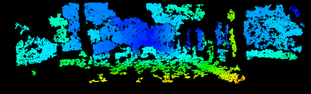}
			\adjincludegraphics[width=\textwidth,trim={{0.1\width} 0 {0.2\width} 0},clip]{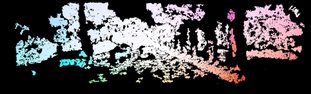}
		\end{subfigure}
		\begin{subfigure}[t]{0.47\columnwidth}
			\adjincludegraphics[width=\textwidth,trim={{0.1\width} 0 {0.2\width} 0},clip]{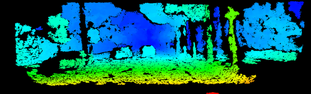}
			\adjincludegraphics[width=\textwidth,trim={{0.1\width} 0 {0.2\width} 0},clip]{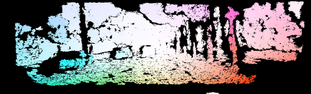}
		\end{subfigure}\\%
		\vspace{0.5mm}%
		\hspace{0.04\columnwidth}
		\begin{subfigure}[t]{0.47\columnwidth}
			\centering
			\small Original Results
		\end{subfigure}
		\begin{subfigure}[t]{0.47\columnwidth}
			\centering
			\small Improved with SDC
		\end{subfigure}		
	\end{center}
	\caption{Our new SDC feature descriptor improves pixel-wise matching in terms of accuracy and density in state-of-the-art algorithms. From top to bottom: Disparity map for ELAS \cite{geiger2010elas} on ETH3D \cite{schops2017multi}, optical flow for CPM \cite{hu2016efficient} on Sintel \cite{butler2012sintel}, and scene flow (disparity and optical flow components) for SFF \cite{schuster2018sceneflowfields} on KITTI \cite{menze2015object}.}
	\label{fig:visual_comparison}
\end{figure}

Applications for driver assistance, robot navigation, autonomous vehicles, and others require a detailed and accurate perception of the environment. Many of these high level computer vision tasks are based on finding pixel-wise correspondences across different images (\eg optical flow or stereo, see \cref{fig:visual_comparison}). 
Robust dense matching of pixel positions under unconstrained conditions typically is a very challenging task for several reasons. Perspective deformations, changing lighting conditions, sensor noise, occlusions, and other effects can change the appearance of corresponding image points drastically. Thus, heuristic descriptors (\eg SIFT \cite{lowe1999sift} or CENSUS \cite{zabih1994census}) can produce very dissimilar descriptors for corresponding image points. A key factor to overcome these issues is the size of context information that is considered by a descriptor. However, increasing the patch size introduces spatial invariance  for state-of-the-art descriptors which results in less accurate matching.
Recently, deep neural networks were shown to produce more robust and expressive features.
These networks rely on best practice design decisions from other domains which results in the use of pooling or other striding layers. Such architectures typically achieve a medium sized receptive field only and reduce the spatial resolution of the resulting feature descriptor. Both properties lower the accuracy of the matching task.

In this paper, we present a deep neural network with a large receptive field that utilizes a novel architecture block to compute highly robust, accurate, dense, and discriminative descriptors for images. To this end, we stack parallel dilated convolutions (SDC).
Our design follows two key observations. First, image patches with low entropy lead to poor descriptors and thus to incorrect matching. This fact strengthens the common belief that a robust descriptor should have a large receptive field to incorporate context knowledge for pixels under difficult visual conditions. Secondly, accurate matching requires a high spatial precision that is lost when applying striding layers which produce coarse, high-level features for deeper layers of the feature network. Our novel architecture block provides a large receptive field with only few trainable parameters while maintaining full spatial resolution.
Overall, our contribution consists of the following:
\begin{itemize}[noitemsep,topsep=1pt,label={\tiny\raisebox{0.75ex}{\textbullet}},leftmargin=*]
\item By stacking multiple, parallel dilated convolutions (SDC), we create a novel neural network block which is beneficial for any dense, pixel-wise prediction task that requires high spatial accuracy.
\item The combination of these blocks to a fully convolutional architecture with a large receptive field that can be used for feature description.
\item Vast sets of experiments to justify our design decisions, to compare to other descriptors, and to demonstrate the accuracy and robustness for scene flow, optical flow, and stereo matching on the well known public data sets KITTI \cite{menze2015object}, MPI Sintel \cite{butler2012sintel}, Middlebury \cite{baker2011database,scharstein2002taxonomy}, HD1K \cite{kondermann2016hci}, and ETH3D \cite{schops2017multi} with our unified network.
\end{itemize}

\section{Related Work} \label{sec:related_work}
A feature descriptor is a vector that represents the characteristics of the associated object in a compact, distinctive manner. It is not to be confused with an interest point (or key point, sometimes feature point) which identifies locations where a feature descriptor would be rather unique. In the context of dense matching, feature descriptors on pixel-level are required. Since single pixels carry only very little information, a region around each pixel is considered for the description. 

Conventional descriptors are often based on image gradients to make them invariant to changes in lighting. A very common descriptor -- SIFT \cite{lowe1999sift} -- computes histograms of gradients in regular grids around the center pixel. Using a multi-scale search and the major orientation of the gradients makes SIFT robust to changes in scale and rotation. However, SIFT was not designed to describe all pixels of an image in a dense manner. The full description is rather slow and sensitive to deformations, occlusions, and motions. Robustness is also a problem for faster hand-crafted feature extractors like SURF \cite{bay2006surf} and DAISY \cite{tola2010daisy}.
Binary descriptors (\eg BRIEF \cite{calonder2010brief}, ORB \cite{rublee2011orb}, or CENSUS \cite{zabih1994census}) are even faster since they are more compact. At the same time, they are less expressive and less distinctive.

To improve robustness, many have applied deep learning for feature extraction on patch-level recently. In \cite{han2015matchnet,zagoruyko2015learning}, features are learned jointly with a decision metric to distinguish corresponding and non-matching image patches with a siamese architecture \cite{chopra2005learning}. For the same reason as L2Net \cite{tian2017l2net}, we do not include a decision network because we want universal features that can be used within any pipeline. The architecture of L2Net \cite{tian2017l2net} avoids pooling layers but requires strided convolution to achieve a medium sized receptive field of $32$ pixels. Additionally, they have experimented with a two-stream design where the input of the second branch is the up-scaled central part of the original patch similar as in \cite{zagoruyko2015learning}. In contrast, our architecture exploits multi-scale information inherently as described in \cref{sec:layer}.

Deep features for the optical flow task were proposed by \cite{bailer2017cnn,gadot2016patchbatch}. PatchBatch introduced batch normalization for patch description for the first time, and \cite{bailer2017cnn} utilized a new thresholded hinge loss. Both architectures consist of several convolutions and pooling layers to obtain considerably large receptive fields. As motivated earlier, our design can easily increase the size of the receptive field without losing the spatial accuracy as it happens during pooling.

For stereo matching, previous work used very light-weight architectures with small receptive fields in favor of speed \cite{zbontar2015computing,luo2016efficient}. For the limited search in stereo matching, the expressiveness of these networks might be sufficient. In contrast, our universal descriptor network for different tasks and domains uses much more context information.

\begin{figure}[t]
	\centering
	\includegraphics[width=1\columnwidth]{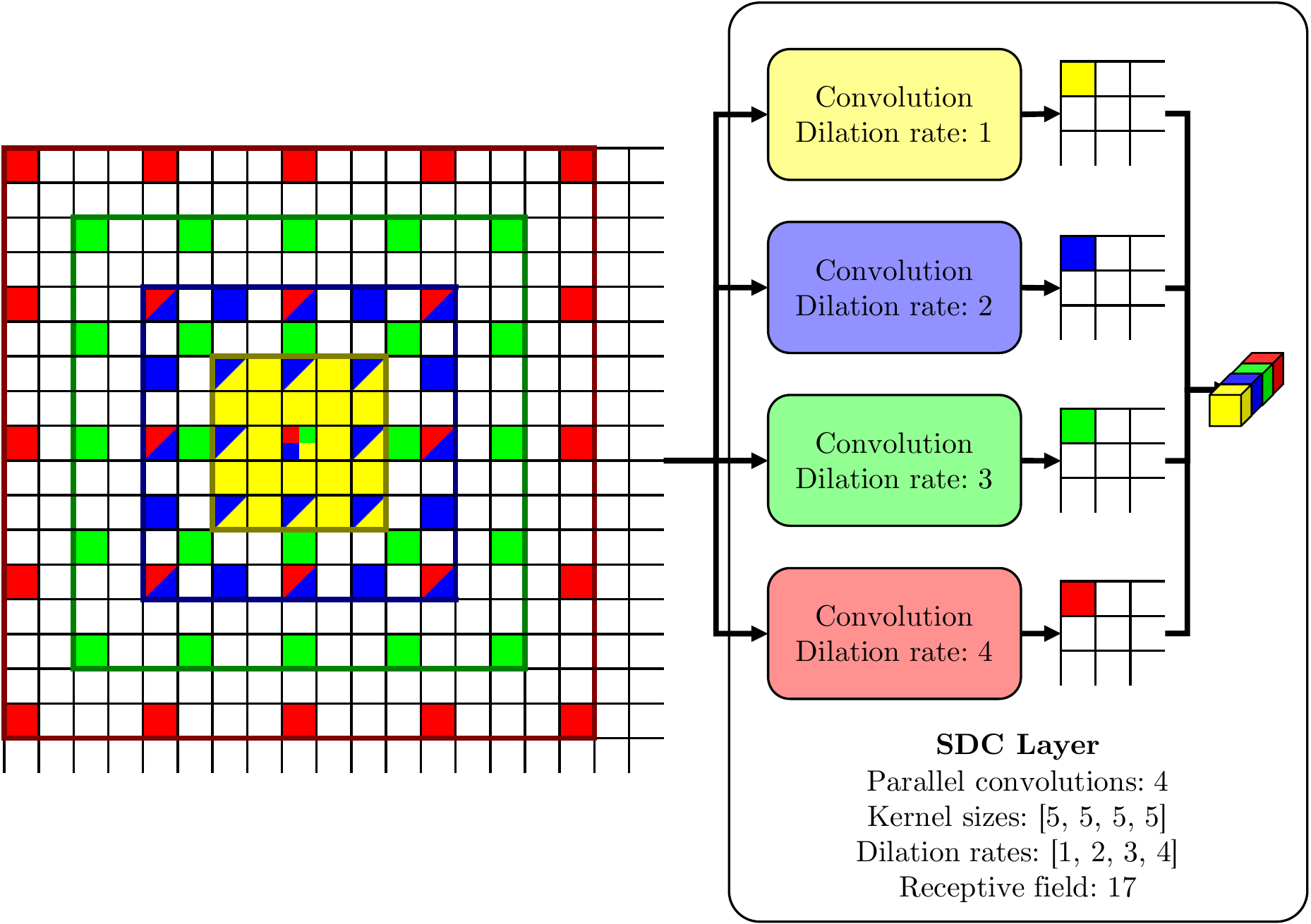}
	\caption{The architecture of a single SDC layer. Our contribution is the combination of parallel convolutions with different dilation rates. The outputs are stacked along the feature dimension to produce a multi-scale response.}
	\label{fig:layer}
\end{figure}

Another concept that is heavily used in our work is atrous or dilated convolution \cite{yu2016multi}. It is a generalization of regular convolution where the kernels are widened by inserting zeros (cf. \cref{fig:layer}). This effectively increases the kernel's perceptive field without adding more parameters or losing spatial details. These advantages were mostly exploited in state-of-the-art semantic segmentation networks \cite{hamaguchi2018effective,hoffer2015deep,wang2018understanding,yu2016multi} by cascading several dilated convolution layers with different dilation factors. Other architectures use dilated convolutions for context aggregation in an end-to-end network after constructing coarse high-level features \cite{guney2016deep}. 
Our novel concept stacks multiple dilated convolutions (SDC) in parallel and combines each output by concatenation to form a single SDC block. This is similar to the Atrous Spatial Pyramid Pooling (ASPP) in \cite{chen2018deeplab} with two major differences. Firstly, we do not sum the parallel results, but stack them. Secondly, our parallel block is not used for feature pooling in a deeper stage of the network, but for feature computation in the first and only stage of our network. That is also why our dilation rates are much smaller in comparison. Another similar combination of dilated convolutions was recently presented in \cite{wei2018revisiting}. They stack a block which is similar to ASPP on top of the DeepLab  \cite{chen2018deeplab} model which boosted performance of object localization significantly. However, \cite{chen2018deeplab,wei2018revisiting} both exploit parallel dilated convolution for semantic context pooling, while we, for the first time, use convolution with different dilation rates to compute multi-scale feature descriptors.

\begin{figure*}[t]
	\centering
	\includegraphics[width=0.95\textwidth]{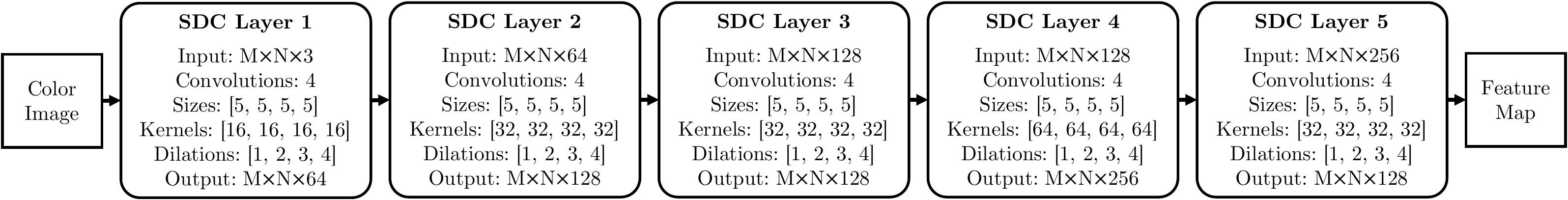}
	\caption{Our SDC feature network. It consists of 5 SDC blocks with varying number of output channels. The final feature vectors are normalized to unit range pixel-wise.}
	\label{fig:network}
\end{figure*}

\section{Feature Network} \label{sec:architecture}
Historically, a large receptive field in convolutional neural networks (CNNs) is primarily obtained by using striding layers. These are typically pooling layers and more recently, pooling is replaced by strided convolution \cite{springenberg2014striving}. Striding layers also improve run time by reducing the size of intermediate representations and introduce some translation invariance. For tasks like image classification, these benefits come at no cost since only a single prediction per image is required. For tasks which require a dense per-pixel prediction, strided layers have the disadvantage of reducing the spatial resolution. This makes pixel-wise prediction overly smooth and less accurate.

The obvious way to obtain a large receptive field without striding is to use larger kernels. The drawbacks of this approach are a drastic increase in run-time and number of parameters which makes such networks slow and prone to overfitting.
This problem can be surpassed by dilated convolution because although the kernels are large, they are sparse (in a regular way). Yet, a sequence of dilated convolutions can introduce gridding effects (different output nodes use disjoint subsets of input nodes) if dilation rates are not selected properly \cite{wang2018understanding}.
As a consequence, we have created a block of stacked dilated convolutions (SDC) in parallel of which the outputs are concatenated. This way, each subsequent layer has full access to previous features of different dilation rates. 

\subsection{SDC Layer} \label{sec:layer}
As others before \cite{chen2018deeplab,li2018csrnet}, we argue that convolution with dilation rate $r$ and stride $r$ is equal to convolution with dilation rate $1$ (no dilation) of sub-sampled input by factor $r$ (no smoothing). Dilated convolution without striding thus produces a sub-scale response at full spatial resolution. This key observation is heavily used by our SDC layer design where we stack the output of convolutions with different dilation rates to produce a multi-scale response (see \cref{fig:layer}). Whereas others apply pooling over multiple scales, we feed the entire multi-scale information to the next layers.

We note that convolution with parallel dilated kernels is similar to convolution with a single larger, sparse kernel (merging the dilated kernels). However, expressiveness is lost where the different dilated kernels overlap (see \cref{fig:layer}). Further, only very few deep learning frameworks support sparse convolution in an efficient way. Nonetheless, an experimental comparison between both designs is provided in the supplementary material.

\subsection{SDC Network} \label{sec:network}
Following the interpretation of dilated convolution of the previous section, we conclude to stack several SDC layers to compute, aggregate, and pass information for multiple scales from end to end. This naturally results in an exponentially growing receptive field but avoids gridding effects because every convolution is fed with the results of every previous convolution of all dilation rates. 

The complete network is illustrated in \cref{fig:network}. We use 5 SDC layers. Each SDC layer applies four parallel convolutions with $5 \times 5$ kernels, the same number of output dimensions, and dilation rates of $1, 2, 3, \text{and } 4$. Exponential Linear Unit (ELU) \cite{clevert2015fast} is used for all activations. We do not use batch normalization because we train with a small batch size (cf. \cref{sec:training}). The SDC layers have $64, 64, 128, 256, \text{and } 128$ output channels respectively. The final feature vector of the last layer is normalized to unit range. Experiments to justify the decision for this design are presented in the supplementary material. Our setup yields a receptive field of $81$ pixels.

Because we do not use any striding, dense image features can be computed in a single forward pass without patch extraction. This makes our design much faster than previous deep descriptors \cite{bailer2017cnn,gadot2016patchbatch,tian2017l2net} during inference.

Our design provides another advantage that can be used within SDC layers: The same kernels are useful for different scales (especially low level vision filters). Thus, it is reasonable to share weights between the parallel convolutions within one SDC block. The only requirement is that the parallel convolutions are of the same shape. By sharing weights, the amount of parameters gets divided by the number of parallel convolutions (factor $4$ in our case). This allows to construct very light-weight feature networks with a comparatively large receptive field. To demonstrate that, we drive network size to an extreme. In our experiments in \cref{sec:experiments}, we train a network with only about $5$~\% of the parameters of our original design that we call \textit{Tiny}. The \textit{Tiny} network has only 4 SDC blocks, each with only 3 parallel dilated convolutions of $3 \times 3$ kernels and dilation rates $1, 2, \text{and } 3$ which share their weights, yielding a receptive field of $25$ pixels.

\subsection{Training Details} \label{sec:training}
Our goal is a universal feature descriptor. Thus, we train a unified feature network on multi-domain data. We use images of the training splits of the following data sets: Scene flow quadtuplets of KITTI 2015 \cite{menze2015object}, optical flow and stereo pairs from MPI Sintel \cite{butler2012sintel},  Middlebury stereo data version 3 \cite{scharstein2002taxonomy}, Middlebury Optical Flow data \cite{baker2011database}, HD1K Benchmark Suite for optical flow \cite{kondermann2016hci}, and the two-view stereo data from ETH3D \cite{schops2017multi}. This is the union of data sets which are used in the Robust Vision Challenge\footnote{\url{www.robustvision.net}} for optical flow and stereo.
We further split 20~\% and 10~\% from the KITTI training set for validation during training and evaluation of our experiments in \cref{sec:experiments} respectively.
Since image sizes, sequence count and lengths vary strongly between data sets, we sample image pairs non-uniformly from each set and then select $100$ patches from the reference image. For each reference patch, we use the ground truth displacement of non-occluded image regions and sample the corresponding patch from the second view. Additionally, we sample a third patch from the second view by altering the ground truth displacement with a random offset to obtain a negative correspondence. All details about the patch sampling along with examples for the sampled triplets can be found in the supplementary material.

\begin{figure}[t]
	\centering
	\includegraphics[width=0.95\columnwidth]{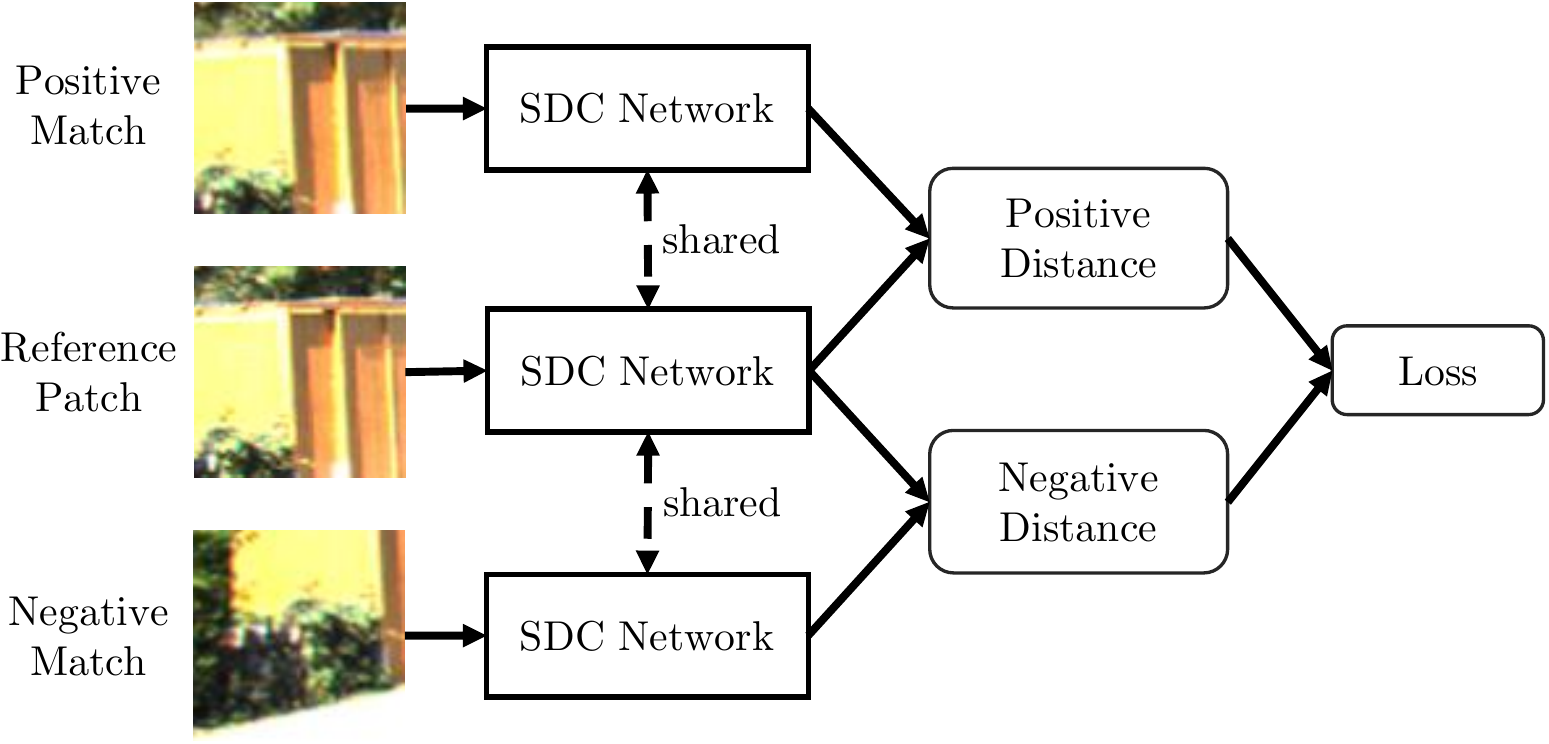}
	\caption{Visualization of the triplet training. For each patch triplet, we compute the loss based on the distance of the feature descriptors for corresponding and non-corresponding patches.}
	\label{fig:training}
\end{figure}

We use a triplet training approach \cite{hoffer2015deep} where we feed the reference patch, the matching patch and the non-matching patch to three of our SDC networks with shared weights. For training stability, we normalize the input by subtracting the mean and dividing by the standard deviation of all training images. As objective function, we choose the thresholded hinge embedding loss of \cite{bailer2017cnn} defined in \cref{eq:loss}.
\begin{equation} \label{eq:loss}
	\begin{split}
	\mathcal{L}\left( r, p, n \right) & = max \left(0, {\lVert f \left( r \right) - f \left( p \right) \rVert}_2^2 - \tau \right) \\
	& + max \left(0, m + \tau - {\lVert f \left( r \right) - f \left( n \right) \rVert}_2^2 \right),
	\end{split}
\end{equation}
where $\{ r, p, n \}$ is the patch triplet, $f$ is the feature transformation of the network, $\tau$ is the threshold, and $m$ is the margin between matching and non-matching features.
We have also experimented with the SoftMax-Triplet loss from \cite{hoffer2015deep} and the SoftPN loss from \cite{balntas2016pn}. Both showed similar performance while being much less stable in training. An overview of the training strategy is given in \cref{fig:training}.

We choose ADAM \cite{kingma2015adam} as optimizer and train with a batch size of $32$ with an initial learning rate of $0.01$ that we exponentially decrease continuously by a power of $0.7$ every $100$ k iterations. We train for $1$ million iterations where convergence saturates, or until overfitting which we rarely observe in any of our experiments. Overfitting is avoided by the random sampling strategy of image pairs and patch triplets which provides many diverse combinations. Photometric data augmentation could not further improve the training process. Instead, we note a small decrease in performance. To speed up training, we crop the input patches and intermediate feature representations to the maximum required size for the respective dilation rate. The complete training of our network takes about $3$ days on a single GeForce GTX 1080.

\section{Experiments} \label{sec:experiments}
We conduct a series of diverse experiments to validate the superior performance of our approach compared to other feature descriptors in \cref{sec:accuracy}. After demonstrating that SDC features outperform heuristic descriptors as well as other neural networks in image patch comparison, we will test our SDC features with different algorithms for different matching tasks on a large number of diverse data sets in \cref{sec:application}. For all experiments, we use a single unified descriptor network. Unlike others \cite{bailer2017cnn,sun2018pwc}, we do not re-train or fine-tune our network on each individual data set.

\begin{table}[t]
	\centering
	\caption{Comparison of the accuracy for representative state-of-the-art descriptors and our SDC design. For learning approaches, we further provide information about receptive field size (RF) in pixels, number of parameters (Size) and accumulated sub-sampling factor due to striding.} \label{tab:accuracy}
	\resizebox{\columnwidth}{!}{\begin{tabular}{c || c || c | c | c }
		\textbf{Network} & \textbf{Accuracy} & \textbf{RF} & \textbf{Size} & \textbf{Factor} \Bstrut\\
		\hline
		SDC (\textbf{Ours}) & \textbf{97.2~\%} & 81 & 1.95 M & \textbf{1}\Tstrut\\
		LargeNet & 96.8~\% & 81 & 22.5 M & \textbf{1} \\
		L2Net \cite{tian2017l2net} & 96.7~\% & 32 & 1.34 M & 4 \\
		Tiny (\textbf{Ours}) & 96.0~\% & 25 & \textbf{0.12 M} & \textbf{1} \\
		PatchBatch \cite{gadot2016patchbatch} & 95.7~\% & 51 & 0.92 M & 8 \\
		DilNet & 95.5~\% & \textbf{96} & 5.43 M & \textbf{1} \\
		2Stream \cite{zagoruyko2015learning} & 92.3~\% & 64 & 2.41 M & 2 \\
		FFCNN \cite{bailer2017cnn} & 90.6~\% & 56 & 4.89 M & 4\Bstrut\\
		\hline
		BRIEF \cite{calonder2010brief} & 93.7~\% & -- & -- & --\Tstrut\\
		DAISY \cite{tola2010daisy} & 92.1~\% & -- & -- & -- \\
		SIFT \cite{lowe1999sift} & 89.0~\% & -- & -- & --\\
	\end{tabular}}
\end{table}

\subsection{Accuracy, Robustness, ROC}  \label{sec:accuracy}
In this section, we compare our SDC descriptor network to other state-of-the-art descriptors.
Representative classical, heuristic descriptors are SIFT \cite{lowe1999sift}, DAISY \cite{tola2010daisy}, and BRIEF \cite{calonder2010brief}. Furthermore, we train the following architectures of previous work that contain striding layers:
\begin{itemize}[noitemsep,topsep=1pt,label={\tiny\raisebox{0.75ex}{\textbullet}},leftmargin=*]
\item \textit{2Stream}: The central-surround network from \cite{zagoruyko2015learning}.
\item \textit{PatchBatch}: The architecture of \cite{gadot2016patchbatch} which utilizes batch normalization.
\item \textit{L2Net}: The basic variant of \cite{tian2017l2net} with only a single stream and without batch normalization, which we found to perform the best among all variants of this network.
\item \textit{FFCNN}: The FlowFieldsCNN architecture \cite{bailer2017cnn}, which showed great improvements over classical descriptors for optical flow estimation.
\end{itemize}
In addition, we design and evaluate two alternative architectures that avoid striding layers. 
\begin{itemize}[noitemsep,topsep=1pt,label={\tiny\raisebox{0.75ex}{\textbullet}},leftmargin=*]
\item \textit{DilNet}: An example of dilated convolution in a sequence: Conv(7,64,1,1)--Conv(7,64,1,2)--Conv(7,128,1,3)--Conv(7,128,1,4)--Conv(7,128,1,3)--Conv(7,256,1,2)--Conv(7,128,1,1). 
\item \textit{LargeNet}: An example for single, large convolutions without dilation: Conv(17,64,1,1)--Conv(17,64,1,1)--Conv(17,128,1,1)--Conv(17,256,1,1)--Conv(17,128,1,1).
\end{itemize}
The four numbers of each convolution layer Conv($k$,$n$,$s$,$d$) describe square kernel size $k$, number of kernels $n$, stride $s$, and dilation rate $d$. Note that \textit{DilNet} and \textit{LargeNet} try to mimic the shape of our SDC network.
More details about each network are given in \cref{tab:accuracy}.

\begin{figure}[t]
	\centering
	\begin{subfigure}[c]{1\columnwidth}
		\includegraphics[width=1\textwidth]{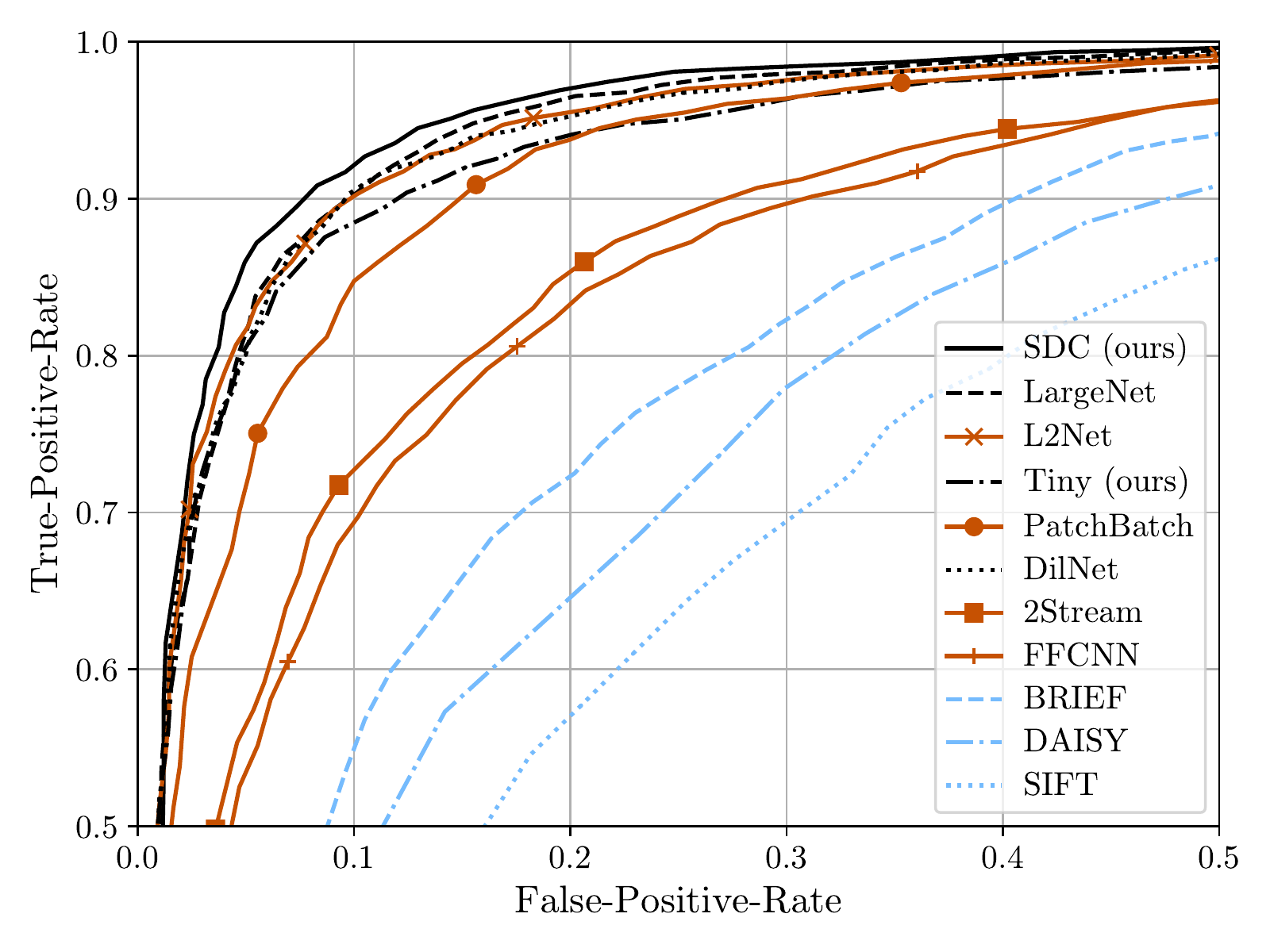}
		\subcaption{ROC curves.}
		\label{fig:plots:roc}
	\end{subfigure}
	\begin{subfigure}[c]{1\columnwidth}
		\includegraphics[width=1\textwidth]{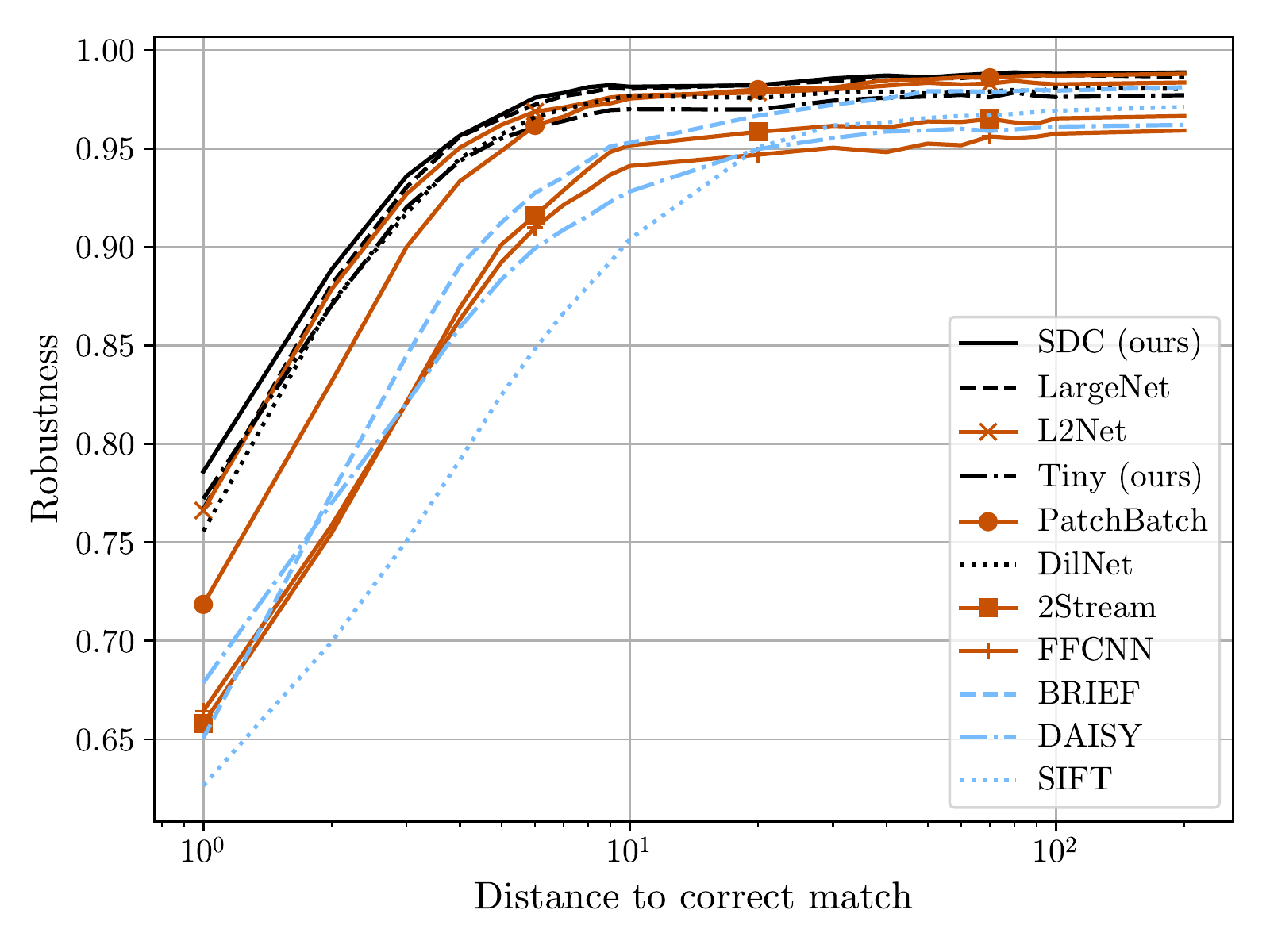}
		\subcaption{Robustness curves.}
		\label{fig:plots:robustness}
	\end{subfigure}
	\caption{In the comparison of ROC and robustness curves, our SDC design outperforms state-of-the-art feature networks and heuristic descriptors.}
	\label{fig:plots}
\end{figure}

First, we evaluate the accuracy of all descriptors. We define accuracy as percentage of correctly distinguished patch triplets, \ie the positive feature distance is smaller than the negative one. Towards that end, we have sampled $2000$ patch triplets from our test images (cf. \cref{sec:training} and the supplementary material). The results are given in \cref{tab:accuracy}.
Our design outperforms all other feature descriptors in terms of accuracy. Our receptive field (RF) is comparatively large, while the network size is comparatively small and we also avoid sub-sampling. Our \textit{Tiny} version is extremely compact without much loss of accuracy. The surprisingly good result of L2Net \cite{tian2017l2net} is worth mentioning, indicating that strided convolution should be preferred over pooling. Also, some of the learning approaches perform worse than the classical descriptors.

We have also computed the Receiver-Operating-Characteristics (ROC) for all descriptors based on the same test triplets. Therefore, we split each triplet into two pairs, a positive and a negative one. True-Positive-Rates over False-Postive-Rates for varying classification thresholds are given in \cref{fig:plots:roc}. Again, our SDC features achieve top performance with a large margin over heuristic descriptors and most neural networks. 

However, matching is not really a classification task. The distance of corresponding descriptors does not matter, as long as it is smaller than these of non-matching descriptors. To take this into account, we have set up a final experiment to show the matching robustness of the descriptors as introduced by \cite{bailer2017cnn}. We have tested each positive corresponding patch pair of our test data against all other correspondences within a certain distance to the correct match. The results are shown in \cref{fig:plots:robustness}. Naturally, the robustness is higher for larger distances to the correct patch. This experiment validates the effectiveness of our design once again. SDC achieves the highest robustness throughout the whole range of displacements. Our top performance is then followed by a dense cluster of other deep descriptors including our \textit{Tiny} variant.
Note the performance of all networks which are explicitly designed to avoid sub-sampling (no strides greater than 1), especially for small offsets.

\subsection{Cross-Task and Cross-Domain Matching}  \label{sec:application}
For the second part of our experiments, we apply our feature descriptor in actual matching tasks. In total, we test 5 algorithms for 3 dense matching tasks with overall 6 data sets. For stereo matching, we evaluate ELAS \cite{geiger2010elas} and SGM \cite{hirschmuller2008SGM} on KITTI \cite{menze2015object}, Middlebury \cite{scharstein2002taxonomy}, and ETH3D \cite{schops2017multi}. CPM \cite{hu2016efficient} and FlowFields++ \cite{schuster2018ffpp} are selected to represent optical flow matching algorithms and are evaluated on KITTI \cite{menze2015object}, Middlebury \cite{baker2011database}, HD1K \cite{kondermann2016hci}, and MPI Sintel \cite{butler2012sintel}. Finally, we test SceneFlowFields (SFF) \cite{schuster2018sceneflowfields} on KITTI \cite{menze2015object}.
Where possible, we evaluate the non-occluded areas (\textit{noc}) and the full image (\textit{all}) separately, because visual matching is only possible in visible regions. On KITTI, these regions are further split into static background (\textit{bg}) and dynamic foreground (\textit{fg}). For the Middlebury stereo data, we evaluate all levels of resolution: Full (\textit{F}), half (\textit{H}), and quarter resolution (\textit{Q}). For Sintel, we consider the more realistic \textit{final} rendering pass only.
We have computed baseline results for the common error metrics average endpoint error (EPE) and the percentage of outliers with an EPE greater than 3 pixels (\textgreater3px) for all data sets. We then change the feature descriptor of every algorithm to our SDC features and repeat the experiment. It is important to note, that we change nothing but the descriptor. For the sake of comparability, we do not fine tune any algorithm, though we expect fine-tuning to improve the results in general.

\paragraph{Stereo Matching.} \label{sec:application:stereo}
ELAS \cite{geiger2010elas} uses first order image gradients for feature description. We use the default parameter set called \textit{MIDDLEBURY} which includes interpolation after consistency check. In addition, we obtain an open source implementation of SGM\footnote{\url{www.github.com/gishi523/semi-global-matching}} which uses the symmetric CENSUS transform \cite{spangenberg2013weighted} of $9 \times 7$ patches as a descriptor.

\begin{table}[t]
	\centering
	\caption{Evaluation of stereo matching algorithms. We compare ELAS \cite{geiger2010elas} and SGM \cite{hirschmuller2008SGM} with the default descriptors and our SDC features on KITTI \cite{menze2015object}, Middlebury \cite{scharstein2002taxonomy}, and ETH3D \cite{schops2017multi}.}
	\label{tab:stereo}
	\resizebox{1\columnwidth}{!}{\begin{tabular}{c c c || c c | c c || c c | c c }
		\multicolumn{3}{c||}{\multirow{3}{*}{Data set}} & \multicolumn{4}{c||}{ELAS \cite{geiger2010elas}} & \multicolumn{4}{c}{SGM \cite{hirschmuller2008SGM}}\\
		& & & \multicolumn{2}{c|}{Original} & \multicolumn{2}{c||}{\textbf{SDC (ours)}} & \multicolumn{2}{c|}{Original} & \multicolumn{2}{c}{\textbf{SDC (ours)}}\\
		& & & \textgreater3px & EPE & \textgreater3px & EPE & \textgreater3px & EPE & \textgreater3px & EPE\Bstrut\\
		\hline
		\multirow{6}{*}{\rotatebox[origin=c]{90}{KITTI}} & \multirow{3}{*}{\rotatebox[origin=c]{90}{\textit{noc}}} & \textit{bg} & 6.56 & 1.30 & \cellcolor{green!100}4.30 & \cellcolor{green!100}1.08 & 4.32 & 1.02 & \cellcolor{green!100}3.44 & \cellcolor{green!25}0.98\Tstrut\\
		& & \textit{fg} & 12.21 & 1.88 & \cellcolor{green!100}8.25 & \cellcolor{green!100}1.41 & 6.46 & 1.15 & \cellcolor{red!100}7.70 & \cellcolor{red!100}1.40 \\
		& & \textit{all} & 7.39 & 1.38 & \cellcolor{green!100}4.88 & \cellcolor{green!100}1.13 & 4.36 & 1.04 & \cellcolor{green!25}4.06 & \cellcolor{white}1.04\Bstrut\\ \hhline{~--||--|--||--|--}
		& \multirow{3}{*}{\rotatebox[origin=c]{90}{\textit{all}}} & \textit{bg} & 7.22 & 1.34 & \cellcolor{green!100}4.86 & \cellcolor{green!100}1.12 & 4.65 & 1.11 & \cellcolor{green!100}3.61 & \cellcolor{green!100}1.00\Tstrut\\
		
		& & \textit{fg} & 14.34 & 2.02 & \cellcolor{green!100}11.28 & \cellcolor{green!100}1.62 & 7.25 & 1.54 & \cellcolor{red!100}8.61 & \cellcolor{red!25}1.68\\
		& & \textit{all} & 8.29 & 1.45 & \cellcolor{green!100}5.83 & \cellcolor{green!100}1.19 & 5.03 & 1.18 & \cellcolor{green!100}4.34 & \cellcolor{green!25}1.10\Bstrut\\
		\hline
		\multirow{6}{*}{\rotatebox[origin=c]{90}{Middlebury}} & \multirow{3}{*}{\rotatebox[origin=c]{90}{\textit{noc}}} & \textit{F} & \color{gray}26.33 & \color{gray}20.42 & \color{gray}22.24 & \color{gray}20.08 & \color{gray}43.92 & \color{gray}44.45 & \color{gray}45.52 & \color{gray}52.41\Tstrut\\
		& & \textit{H} & 16.85 & 4.44 & \cellcolor{green!100}12.03 & \cellcolor{green!100}3.42 & 15.93 & 6.12 & \cellcolor{green!100}13.37 & \cellcolor{red!100}6.98\\
		& & \textit{Q} & 11.62 & 2.03 & \cellcolor{green!100}10.12 & \cellcolor{green!25}1.91 & 10.43 & 1.81 & \cellcolor{green!100}8.80 & \cellcolor{green!25}1.75\Bstrut\\ \hhline{~--||--|--||--|--}
		& \multirow{3}{*}{\rotatebox[origin=c]{90}{\textit{all}}} & \textit{F} & \color{gray}29.87 & \color{gray}22.47 & \color{gray}26.22 & \color{gray}22.16 & \color{gray}47.28 & \color{gray}47.56 & \color{gray}48.09 & \color{gray}53.50\Tstrut\\
		& & \textit{H} & 21.02 & 6.03 & \cellcolor{green!100}16.97 & \cellcolor{green!100}5.08 & 19.71 & 7.64 & \cellcolor{green!100}16.73 & \cellcolor{red!25}8.26\\
		& & \textit{Q} & 15.91 & 2.91 & \cellcolor{green!25}15.19 & \cellcolor{green!25}2.86 & 14.77 & 2.74 & \cellcolor{green!100}12.26 & \cellcolor{green!100}2.46\Bstrut\\
		\hline
		\multirow{3}{*}{\rotatebox[origin=c]{90}{ETH3D}} & \multicolumn{2}{c||}{\textit{noc}} & 6.03 & 0.98 & \cellcolor{green!100}2.17 & \cellcolor{green!100}0.60 & 2.83 & 0.65 & \cellcolor{red!100}3.11 & \cellcolor{red!100}0.75 \Tstrut\\ 
		& \multicolumn{2}{c||}{\textit{occ}} & 17.68 & 2.14 & \cellcolor{green!100}12.99 & \cellcolor{green!100}1.64 & 6.40 & 1.36 & \cellcolor{green!100}4.81 & \cellcolor{green!100}1.11 \\
		& \multicolumn{2}{c||}{\textit{all}} & 6.50 & 1.02 & \cellcolor{green!100}2.61 & \cellcolor{green!100}0.64 & 3.62 & 0.81 & \cellcolor{green!25}3.49 & \cellcolor{red!25}0.83 \\
	\end{tabular}}
\end{table}

\begin{table*}[t]
	\centering
	\caption{Optical flow evalation with FlowFields++ \cite{schuster2018ffpp}. We compare SIFT \cite{lowe1999sift} to our SDC features on KITTI \cite{menze2015object}, Sintel \cite{butler2012sintel}, Middlebury \cite{baker2011database}, and HD1K \cite{kondermann2016hci}. Results for dense matching, after consistency check, and after interpolation are shown.}
	\label{tab:ffpp}
	\resizebox{0.75\width}{!}{\begin{tabular}{c c c || c c | c c | c c c | c c c | c c | c c }
		\multicolumn{3}{c||}{\multirow{3}{*}{Data set}} & \multicolumn{4}{c|}{Matching} & \multicolumn{6}{c|}{Filtered} & \multicolumn{4}{c}{Interpolated}\\
		& & & \multicolumn{2}{c|}{SIFT \cite{lowe1999sift}} & \multicolumn{2}{c|}{\textbf{SDC (ours)}}& \multicolumn{3}{c|}{SIFT \cite{lowe1999sift}} & \multicolumn{3}{c|}{\textbf{SDC (ours)}} & \multicolumn{2}{c|}{SIFT \cite{lowe1999sift}} & \multicolumn{2}{c}{\textbf{SDC (ours)}}\\
		& & & \textgreater3px & EPE & \textgreater3px & EPE & \textgreater3px & EPE & Density & \textgreater3px & EPE & Density & \textgreater3px & EPE & \textgreater3px & EPE\Bstrut\\
		\hline
		\multirow{6}{*}{\rotatebox[origin=c]{90}{KITTI}} & \multirow{3}{*}{\rotatebox[origin=c]{90}{\textit{noc}}} & \textit{bg} & 23.22 &  12.07 & \cellcolor{green!100}15.25 & \cellcolor{green!100}6.56 & 8.04 & 1.89 & -- & \cellcolor{green!100}6.91 & \cellcolor{red!25}2.00 & -- & 9.56 & 3.08 & \cellcolor{green!100}8.52 & \cellcolor{green!25}2.97\Tstrut\\
		& & \textit{fg} & 27.61 & 14.47 & \cellcolor{green!100}16.90 & \cellcolor{green!100}4.45 & 10.31 & 2.11 & -- & \cellcolor{green!100}9.10 & \cellcolor{green!25}2.01 & -- & 6.13 & 1.97 & \cellcolor{red!100}8.99 & \cellcolor{red!100}2.57\\
		& & \textit{all} & 23.98 & 12.48 & \cellcolor{green!100}15.53 & \cellcolor{green!100}6.19 & 8.39 & 1.92 & 73.3~\% & \cellcolor{green!100}7.27 & \cellcolor{red!25}2.00 & \cellcolor{green!100}86.1~\% & 8.97 & 2.89 & \cellcolor{green!25}8.60 & \cellcolor{red!25}2.90\Bstrut\\ \hhline{~--||--|--|---|---|--|--}
		& \multirow{3}{*}{\rotatebox[origin=c]{90}{\textit{all}}} & \textit{bg} & 35.96 & 53.20 & \cellcolor{green!100}29.19 & \cellcolor{green!100}39.42 & 9.02 & 3.09 & -- & \cellcolor{green!25}8.30 & \cellcolor{red!100}3.62 & -- & 19.13 & 9.45 & \cellcolor{green!100}17.19 & \cellcolor{green!25}9.13\Tstrut\\		
		& & \textit{fg} & 29.17 & 21.81 & \cellcolor{green!100}18.64 & \cellcolor{red!100}24.15 & 10.32 & 2.11 & -- & \cellcolor{green!100}9.10 & \cellcolor{green!25}2.01 & -- & 6.46 & 2.14 & \cellcolor{red!100}9.19 & \cellcolor{red!100}2.71\\
		& & \textit{all} & 34.93 & 48.45 & \cellcolor{green!100}27.59 & \cellcolor{green!100}37.11 & 9.22 & 2.94 & 63.3~\% & \cellcolor{green!25}8.43 & \cellcolor{red!100}3.36 & \cellcolor{green!100}74.7~\% & 17.21 & 8.34 & \cellcolor{green!100}15.98 & \cellcolor{green!25}8.16\Bstrut\\
		\hline
		\multirow{3}{*}{\rotatebox[origin=c]{90}{Sintel}} & \multicolumn{2}{c||}{\textit{noc}} & 16.21 & 9.31 & \cellcolor{green!100}10.15 & \cellcolor{green!100}5.17 & 4.15 & 1.00 & -- & \cellcolor{green!100}3.61 & \cellcolor{green!25}0.96 & -- & 6.35 & 2.34 & \cellcolor{green!25}6.10 & \cellcolor{green!25}2.18\Tstrut\\ 
		& \multicolumn{2}{c||}{\textit{occ}} & 83.18 & 120.78 & \cellcolor{green!25}78.85 & \cellcolor{green!100}89.72 & 42.59 & 10.10 & -- & \cellcolor{red!25}43.81 & \cellcolor{red!25}10.71 & -- & 45.04 & 22.26 & \cellcolor{red!25}46.80 & \cellcolor{green!25}21.33\\
		& \multicolumn{2}{c||}{\textit{all}} & 21.88 & 18.75 & \cellcolor{green!100}15.97 & \cellcolor{green!100}12.33 & 5.15 & 1.23 & 75.7~\% & \cellcolor{green!25}4.82 & \cellcolor{red!25}1.25 & \cellcolor{green!100}84.4~\% & 9.62 & 4.03 & \cellcolor{green!25}9.55 & \cellcolor{green!25}3.80\Bstrut\\
		\hline
		\multicolumn{3}{c||}{Middlebury} & 5.47 & 1.21 & \cellcolor{green!100}3.79 & \cellcolor{green!100}0.76 & 2.24 & 0.51 & 93.1~\% & \cellcolor{red!25}2.26 & \cellcolor{green!25}0.49 & \cellcolor{green!25}96.9~\% & 1.69 & 0.28 & \cellcolor{red!25}1.79 & \cellcolor{red!25}0.30\Tstrut\Bstrut\\
		\hline
		\multicolumn{3}{c||}{HD1K} & 15.52 & 12.99 & \cellcolor{green!100}7.76 & \cellcolor{green!100}7.48 & 5.64 & 1.16 & 82.6~\% & \cellcolor{green!100}4.10 & \cellcolor{green!100}1.02 & \cellcolor{green!100}94.4~\% & 4.34 & 0.96 & \cellcolor{red!25}4.62 & \cellcolor{red!100}1.31\Tstrut\\
	\end{tabular}}
\end{table*}

\begin{table}[t]
	\centering
	\caption{Evaluation of optical flow matching with CPM \cite{hu2016efficient}. We compare SIFT \cite{lowe1999sift} and our SDC features on KITTI \cite{menze2015object}, Sintel \cite{butler2012sintel}, Middlebury \cite{baker2011database}, and HD1K \cite{kondermann2016hci}.}
	\label{tab:cpm}
	\resizebox{0.72\width}{!}{\begin{tabular}{c c c || c c c | c c c }
		\multicolumn{3}{c||}{\multirow{2}{*}{Data set}} & \multicolumn{3}{c|}{SIFT \cite{lowe1999sift}} & \multicolumn{3}{c}{\textbf{SDC (ours)}}\\
		& & & \textgreater3px & EPE & Density & \textgreater3px & EPE & Density\Bstrut\\
		\hline
		\multirow{6}{*}{\rotatebox[origin=c]{90}{KITTI}} & \multirow{3}{*}{\rotatebox[origin=c]{90}{\textit{noc}}} & \textit{bg} & 10.69 & 2.17 & -- & \cellcolor{green!100}8.37 & \cellcolor{red!25}2.30 & --\Tstrut\\
		& & \textit{fg} & 12.67 & 2.40 & -- & \cellcolor{green!100}9.96 & \cellcolor{green!100}2.14 & --\\
		& & \textit{all} & 11.26 & 2.21 & 7.88~\% & \cellcolor{green!100}8.64 & \cellcolor{red!25}2.30 & \cellcolor{green!100}9.83~\%\Bstrut\\ \hhline{~--||---|---}
		& \multirow{3}{*}{\rotatebox[origin=c]{90}{\textit{all}}} & \textit{bg} & 11.71 & 3.28 & -- & \cellcolor{green!100}9.48 & \cellcolor{red!100}3.79 & --\Tstrut\\		
		& & \textit{fg} & 12.67 & 2.40 & -- & \cellcolor{green!100}9.96 & \cellcolor{green!100}2.14 & --\\
		& & \textit{all} & 11.87 & 3.13 & 6.79~\% & \cellcolor{green!100}9.56 & \cellcolor{red!100}3.51 & \cellcolor{green!100}8.50~\%\Bstrut\\
		\hline
		\multirow{3}{*}{\rotatebox[origin=c]{90}{Sintel}} & \multicolumn{2}{c||}{\textit{noc}} & 4.33 & 1.06 & -- & \cellcolor{red!25}4.64 & \cellcolor{red!100}1.18 & --\\ 
		& \multicolumn{2}{c||}{\textit{occ}} & 45.03 & 10.49 & -- & \cellcolor{red!25}49.52 & \cellcolor{red!100}12.56 & --\\
		& \multicolumn{2}{c||}{\textit{all}} & 5.30 & 1.28 & 8.93~\% & \cellcolor{red!100}5.90 & \cellcolor{red!100}1.50 & \cellcolor{green!100}9.52~\%\Bstrut\\
		\hline
		\multicolumn{3}{c||}{Middlebury} & 4.11 & 0.79 & 10.10~\% & \cellcolor{green!100}2.57 & \cellcolor{green!100}0.66 & \cellcolor{green!25}10.49~\%\Tstrut\Bstrut\\
		\hline
		\multicolumn{3}{c||}{HD1K} & 5.85 & 1.29 & 9.80~\% & \cellcolor{green!100}4.46 & \cellcolor{green!25}1.17 & \cellcolor{green!25}10.54~\%\Tstrut\\
	\end{tabular}}
\end{table}

\begin{figure*}[t]
	\begin{center}
		\begin{subfigure}[c]{0.02\textwidth}
			\rotatebox[origin=c]{90}{\small Image}
		\end{subfigure}
		\begin{subfigure}[c]{0.3312\textwidth}
			\includegraphics[width=\textwidth]{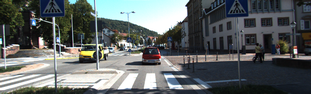}
		\end{subfigure}
		\begin{subfigure}[c]{0.2349\textwidth}
			\includegraphics[width=\textwidth]{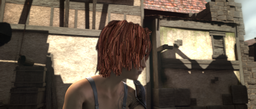}
		\end{subfigure}
		\begin{subfigure}[c]{0.1105\textwidth}
			\includegraphics[width=\textwidth]{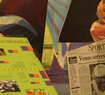}
		\end{subfigure}
		\begin{subfigure}[c]{0.2370\textwidth}
			\includegraphics[width=\textwidth]{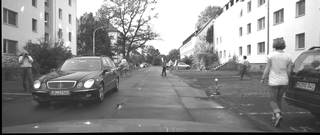}
		\end{subfigure}\\%
		\vspace{0.5mm}%
		\begin{subfigure}[c]{0.02\textwidth}
			\rotatebox[origin=c]{90}{\small Original}
		\end{subfigure}
		\begin{subfigure}[c]{0.3312\textwidth}
			\includegraphics[width=\textwidth]{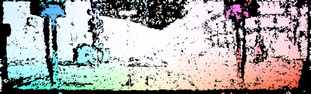}
		\end{subfigure}
		\begin{subfigure}[c]{0.2349\textwidth}
			\includegraphics[width=\textwidth]{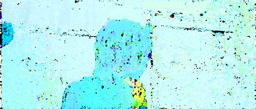}
		\end{subfigure}
		\begin{subfigure}[c]{0.1105\textwidth}
			\includegraphics[width=\textwidth]{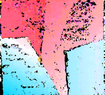}
		\end{subfigure}
		\begin{subfigure}[c]{0.2370\textwidth}
			\includegraphics[width=\textwidth]{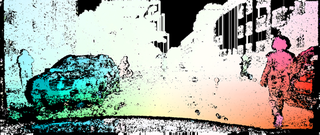}
		\end{subfigure}\\%
		\vspace{0.5mm}%
		\begin{subfigure}[c]{0.02\textwidth}
			\rotatebox[origin=c]{90}{\small with SDC}
		\end{subfigure}
		\begin{subfigure}[c]{0.3312\textwidth}
			\includegraphics[width=\textwidth]{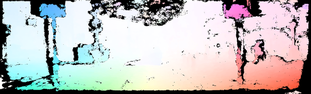}
		\end{subfigure}
		\begin{subfigure}[c]{0.2349\textwidth}
			\includegraphics[width=\textwidth]{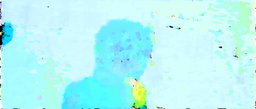}
		\end{subfigure}
		\begin{subfigure}[c]{0.1105\textwidth}
			\includegraphics[width=\textwidth]{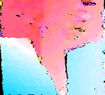}
		\end{subfigure}
		\begin{subfigure}[c]{0.2370\textwidth}
			\includegraphics[width=\textwidth]{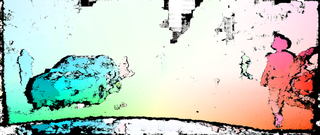}
		\end{subfigure}\\%
		\vspace{0.5mm}%
		\hspace{0.02\textwidth}
		\begin{subfigure}[c]{0.3312\textwidth}
			\centering
			\small KITTI \cite{menze2015object}
		\end{subfigure}
		\begin{subfigure}[c]{0.2299\textwidth}
			\centering
			\small Sintel \cite{butler2012sintel}
		\end{subfigure}
		\begin{subfigure}[c]{0.12\textwidth}
			\centering
			\small Middlebury \cite{baker2011database}
		\end{subfigure}
		\begin{subfigure}[c]{0.2320\textwidth}
			\centering
			\small HD1K \cite{kondermann2016hci}
		\end{subfigure}
	\end{center}
	\vspace{-2.5mm}
	\caption{Exemplary visual comparison of filtered optical flow from FF++ \cite{schuster2018ffpp} on four different data sets. The second row shows results for the original method, while the bottom row shows results after changing the feature descriptor to SDC. Note that all parameters are the same for both experiments. Quantitative evaluation on full data sets is provided in \cref{tab:ffpp}.}
	\label{fig:ffpp}
	\vspace{-4.5mm}
\end{figure*}

\begin{table*}[t]
	\centering
	\caption{Results for scene flow estimation. SceneFlowFields \cite{schuster2018sceneflowfields} with SIFTFlow \cite{liu2011siftflow} features and our SDC features are compared on the KITTI Scene Flow Benchmark \cite{menze2015object}. The densities after filtering increase from 43.6~\% to 67.0~\% in \textit{noc} and from 36.4~\% to 56.0~\% in \textit{all} regions when using SDC features.}
	\label{tab:sff}
	\resizebox{0.72\width}{!}{\begin{tabular}{c c c || c c | c c | c c | c c | c c | c c | c c | c c }
		\multicolumn{3}{c||}{\multirow{3}{*}{Data}} & \multicolumn{4}{c|}{Matching} & \multicolumn{4}{c|}{Filtered} & \multicolumn{4}{c|}{Interpolated} & \multicolumn{4}{c}{Ego-motion Refinement}\\
		& & & \multicolumn{2}{c|}{SIFTFlow} & \multicolumn{2}{c|}{\bf SDC (ours)} & \multicolumn{2}{c|}{SIFTFlow} & \multicolumn{2}{c|}{\bf SDC (ours)} & \multicolumn{2}{c|}{SIFTFlow} & \multicolumn{2}{c|}{\bf SDC (ours)} & \multicolumn{2}{c|}{SIFTFlow} & \multicolumn{2}{c}{\bf SDC (ours)}\\
		& & & \textgreater3px & EPE & \textgreater3px & EPE & \textgreater3px & EPE & \textgreater3px & EPE & \textgreater3px & EPE & \textgreater3px & EPE & \textgreater3px & EPE & \textgreater3px & EPE\Bstrut\\
		\hline
		\multirow{6}{*}{\rotatebox[origin=c]{90}{D1}} & \multirow{3}{*}{\rotatebox[origin=c]{90}{\textit{noc}}} & \textit{bg} & 9.84 & 2.00 & \cellcolor{green!100}4.69 & \cellcolor{green!100}1.10 & 2.29 & 0.85 & \cellcolor{green!100}1.39 & \cellcolor{green!100}0.68 & 4.94 & 1.04 & \cellcolor{green!100}4.26 & \cellcolor{green!100}0.93 & -- & -- & -- & --\Tstrut\\
		& & \textit{fg} & 16.23 & 2.64 & \cellcolor{green!100}9.72 & \cellcolor{green!100}1.89 & 2.76 & 0.80 & \cellcolor{red!25}2.81 & \cellcolor{green!25}0.78 & 7.85 & 1.33 & \cellcolor{green!25}7.59 & \cellcolor{green!100}1.17 & -- & -- & -- & --\\
		& & \textit{all} & 10.91 & 2.11 & \cellcolor{green!100}5.53 & \cellcolor{green!100}1.23 & 2.37 & 0.84 & \cellcolor{green!100}1.60 & \cellcolor{green!100}0.69 & 5.43 & 1.09 & \cellcolor{green!100}4.82 & \cellcolor{green!100}0.97 & -- & -- & -- & --\Bstrut\\
		\hhline{~--||--|--|--|--|--|--|--|--}
		& \multirow{3}{*}{\rotatebox[origin=c]{90}{\textit{all}}} & \textit{bg} & 11.62 & 4.93 & \cellcolor{green!100}6.53 & \cellcolor{red!100}5.63 & 2.30 & 0.85 & \cellcolor{green!100}1.40 & \cellcolor{green!100}0.68 & 5.33 & 1.13 & \cellcolor{green!100}4.58 & \cellcolor{green!100}1.02 & -- & -- & -- & --\Tstrut\\
		& & \textit{fg} & 20.64 & 15.64 & \cellcolor{green!100}14.51 & \cellcolor{green!100}10.08 & 2.76 & 0.80 & \cellcolor{red!25}2.82 & \cellcolor{green!25}0.78 & 7.78 & 1.33 & \cellcolor{red!25}8.20 & \cellcolor{green!25}1.26 & -- & -- & -- & --\\
		& & \textit{all} & 12.99 & 6.55 & \cellcolor{green!100}7.59 & \cellcolor{green!25}6.31 & 2.38 & 0.84 & \cellcolor{green!100}1.61 & \cellcolor{green!100}0.69 & 5.71 & 1.16 & \cellcolor{green!100}5.13 & \cellcolor{green!25}1.06 & -- & -- & -- & --\Bstrut\\
		\hline
		\multirow{6}{*}{\rotatebox[origin=c]{90}{D2}} & \multirow{3}{*}{\rotatebox[origin=c]{90}{\textit{noc}}} & \textit{bg} & 17.49 & 2.82 & \cellcolor{green!100}10.49 & \cellcolor{green!100}1.80 & 2.74 & 0.92 & \cellcolor{green!100}1.94 & \cellcolor{green!100}0.80 & 12.05 & 2.24 & \cellcolor{green!100}7.64 & \cellcolor{green!100}1.35 & 6.89 & 1.47 & \cellcolor{green!100}6.12 & \cellcolor{green!100}1.17\Tstrut\\
		& & \textit{fg} & 16.65 & 2.88 & \cellcolor{green!100}11.41 & \cellcolor{green!100}1.86 & 2.75 & 0.88 & \cellcolor{red!25}2.88 & \cellcolor{white!100}0.88 & 9.91 & 1.69 & \cellcolor{green!100}8.48 & \cellcolor{green!100}1.38 & 10.33 & 1.62 & \cellcolor{green!100}8.61 & \cellcolor{green!100}1.43\\
		& & \textit{all} & 17.35 & 2.83 & \cellcolor{green!100}10.65 & \cellcolor{green!100}1.81 & 2.74 & 0.91 & \cellcolor{green!100}2.08 & \cellcolor{green!100}0.81 & 11.69 & 2.15 & \cellcolor{green!100}7.78 & \cellcolor{green!100}1.36 & 7.47 & 1.49 & \cellcolor{green!100}6.54 & \cellcolor{green!100}1.21\Bstrut\\
		\hhline{~--||--|--|--|--|--|--|--|--}
		& \multirow{3}{*}{\rotatebox[origin=c]{90}{\textit{all}}} & \textit{bg} & 31.38 & 8.47 & \cellcolor{green!100}25.54 & \cellcolor{red!25}8.71 & 2.83 & 0.94 & \cellcolor{green!100}2.11 & \cellcolor{green!100}0.84 & 18.11 & 3.30 & \cellcolor{green!100}12.97 & \cellcolor{green!100}2.20 & 8.80 & 1.82 & \cellcolor{green!25}8.74 & \cellcolor{green!100}1.61\Tstrut\\
		& & \textit{fg} & 20.80 & 5.20 & \cellcolor{green!100}15.91 & \cellcolor{red!25}5.54 & 2.75 & 0.88 & \cellcolor{red!25}2.88 & \cellcolor{white!100}0.88 & 9.85 & 1.68 & \cellcolor{red!25}10.70 & \cellcolor{green!25}1.51 & 10.24 & 1.61 & \cellcolor{red!25}10.82 & \cellcolor{green!25}1.57\\
		& & \textit{all} & 29.61 & 7.97 & \cellcolor{green!100}24.09 & \cellcolor{red!25}8.23 & 2.82 & 0.93 & \cellcolor{green!100}2.23 & \cellcolor{green!25}0.85 & 16.86 & 3.06 & \cellcolor{green!100}12.63 & \cellcolor{green!100}2.09 & 9.02 & 1.79 & \cellcolor{red!25}9.05 & \cellcolor{green!100}1.60\Bstrut\\
		\hline
		\multirow{6}{*}{\rotatebox[origin=c]{90}{Fl}} & \multirow{3}{*}{\rotatebox[origin=c]{90}{\textit{noc}}} & \textit{bg} & 22.95 & 9.07 & \cellcolor{green!100}13.25 & \cellcolor{green!100}5.10 & 2.34 & 0.85 & \cellcolor{red!25}2.55 & \cellcolor{red!100}0.95 & 17.77 & 6.65 & \cellcolor{green!100}10.18 & \cellcolor{green!100}2.52 & 9.42 & 2.27 & \cellcolor{green!100}8.10 & \cellcolor{green!100}2.02\Tstrut\\
		& & \textit{fg} & 25.40 & 7.06 & \cellcolor{green!100}14.42 & \cellcolor{green!100}4.34 & 2.14 & 1.05 & \cellcolor{red!100}3.00 & \cellcolor{red!25}1.19 & 11.48 & 2.73 & \cellcolor{green!100}7.52 & \cellcolor{green!100}1.92 & 13.05 & 3.42 & \cellcolor{green!100}9.07 & \cellcolor{green!100}2.52\\
		& & \textit{all} & 23.36 & 8.74 & \cellcolor{green!100}13.44 & \cellcolor{green!100}4.97 & 2.31 & 0.88 & \cellcolor{red!100}2.61 & \cellcolor{red!100}0.98 & 16.72 & 5.99 & \cellcolor{green!100}9.73 & \cellcolor{green!100}2.42 & 10.03 & 2.46 & \cellcolor{green!100}8.26 & \cellcolor{green!100}2.11\Bstrut\\
		\hhline{~--||--|--|--|--|--|--|--|--}
		& \multirow{3}{*}{\rotatebox[origin=c]{90}{\textit{all}}} & \textit{bg} & 36.68 & 45.26 & \cellcolor{green!100}28.36 & \cellcolor{green!100}38.11 & 2.43 & 0.97 & \cellcolor{red!100}2.71 & \cellcolor{red!100}1.11 & 26.84 & 13.42 & \cellcolor{green!100}18.31 & \cellcolor{green!100}7.44 & 13.04 & 4.71 & \cellcolor{green!100}11.73 & \cellcolor{red!25}5.00\Tstrut\\
		& & \textit{fg} & 29.70 & 17.64 & \cellcolor{green!100}17.60 & \cellcolor{green!100}8.20 & 2.14 & 1.05 & \cellcolor{red!100}3.00 & \cellcolor{red!100}1.19 & 12.52 & 2.83 & \cellcolor{green!100}8.88 & \cellcolor{green!100}2.12 & 13.97 & 3.47 & \cellcolor{green!100}10.31 & \cellcolor{green!100}2.68\\
		& & \textit{all} & 35.47 & 41.08 & \cellcolor{green!100}26.73 & \cellcolor{green!100}33.58 & 2.38 & 0.99 & \cellcolor{red!100}2.76 & \cellcolor{red!100}1.12 & 24.68 & 11.82 & \cellcolor{green!100}16.89 & \cellcolor{green!100}6.64 & 13.18 & 4.52 & \cellcolor{green!100}11.51 & \cellcolor{red!25}4.65\Bstrut\\
		\hline
		\multirow{6}{*}{\rotatebox[origin=c]{90}{SF}} & \multirow{3}{*}{\rotatebox[origin=c]{90}{\textit{noc}}} & \textit{bg} & 29.99 & -- & \cellcolor{green!100}17.65 & -- & 4.51 & -- & \cellcolor{green!100}3.83 & -- & 20.38 & -- & \cellcolor{green!100}12.52 & -- & 11.24 & -- & \cellcolor{green!100}9.46 & --\Tstrut\\
		& & \textit{fg} & 34.97 & -- & \cellcolor{green!100}22.03 & -- & 5.00 & -- & \cellcolor{red!25}5.39 & -- & 16.25 & -- & \cellcolor{green!100}13.72 & -- & 17.48 & -- & \cellcolor{green!100}15.03 & -- \\
		& & \textit{all} & 30.82 & -- & \cellcolor{green!100}17.65 & -- & 4.59 & -- & \cellcolor{green!100}3.83 & -- & 19.69 & -- & \cellcolor{green!100}12.72 & -- & 12.28 & -- & \cellcolor{green!100}10.39 & --\Bstrut\\
		\hhline{~--||--|--|--|--|--|--|--|--}
		& \multirow{3}{*}{\rotatebox[origin=c]{90}{\textit{all}}} & \textit{bg} & 42.68 & -- & \cellcolor{green!100}31.19 & -- & 4.61 & -- & \cellcolor{green!100}3.73 & -- & 26.84 & -- & \cellcolor{green!100}20.46 & -- & 14.68 & -- & \cellcolor{green!100}12.97 & --\Tstrut\\
		& & \textit{fg} & 40.04 & -- & \cellcolor{green!100}28.09 & -- & 5.00 & -- & \cellcolor{red!25}5.40 & -- & 17.36 & -- & \cellcolor{green!25}16.42 & -- & 18.49 & -- & \cellcolor{green!25}17.62 & --\\
		& & \textit{all} & 42.28 & -- & \cellcolor{green!100}31.19 & -- & 4.67 & -- & \cellcolor{green!100}3.98 & -- & 27.43 & -- & \cellcolor{green!100}19.85 & -- & 15.26 & -- & \cellcolor{green!100}13.67 & --\\
	\end{tabular}}
\end{table*}

Results for both algorithms on all stereo data sets are given in \cref{tab:stereo}. Green color indicates where our features outperform the baseline; decrease in accuracy is marked in red. In case of ELAS \cite{geiger2010elas}, the impact of SDC features is advantageous in all cases, and even significant most of the time. SGM \cite{hirschmuller2008SGM} shows a couple of negative test cases. First of all, the full resolution (\textit{F}) images of Middlebury \cite{scharstein2002taxonomy} which produce bad results for both descriptors on both data sets, since the default parameters of ELAS \cite{geiger2010elas} and SGM \cite{hirschmuller2008SGM} are not adjusted to the maximum possible disparity of that resoltuion. This might also apply to the half resolution images (\textit{H}) to some extend. As a consequence, this data should not be considered in the comparison. Then there is the foreground regions of KITTI \cite{menze2015object}, where our deep features perform slightly worse than CENSUS. This might be, because foreground regions are underrepresented in the data set, and thus in the randomly sampled training patches. Lastly, the non-occluded areas of ETH3D \cite{schops2017multi} show minimally higher errors for our features. However, the large receptive field of SDC features can compensate for that in occluded regions to improve the overall results. In summary, SDC features improve dense stereo matching for both algorithms on all data sets.

\paragraph{Optical Flow Correspondences.} \label{sec:application:opticalflow}
CPM \cite{hu2016efficient} computes sparse matches in non-overlapping $3 \times 3$ blocks that can be used for interpolation with EPICFlow \cite{revaud2015epic} or RICFlow \cite{hu2017robust}. The original feature descriptor is SIFT \cite{lowe1999sift}. We evaluate the generated matches of this algorithm in \cref{tab:cpm}. FlowFields++ (FF++) \cite{schuster2018ffpp} performs dense matching, followed by a consistency check and interpolation with RICFlow \cite{hu2017robust}. We compare the results between the originally used SIFT features \cite{lowe1999sift} and our SDC features after each of these 3 steps in \cref{tab:ffpp}. For the filtered results after the consistency check, we also give the density as percentage of covered ground truth pixels. Visual examples are given in \cref{fig:ffpp}.

In some cases, both algorithms show a slight increase in endpoint error for the complete KITTI data (\textit{all}) when used with our SDC features. This is most likely due to the fact, that the KITTI \textit{noc} data excludes the out-of-bounds motions only, not the real occlusions. A higher endpoint error in the occluded areas is actually an advantage, because it makes  outlier filtering during consistency check easier. In fact, EPE and outliers are better for KITTI-\textit{all}-\textit{fg} for FF++ after filtering (see \cref{tab:ffpp}).
Also, it is important to note that the filtered matches with SDC are much denser for both algorithms (cf. \cref{fig:ffpp}). Dense, well distributed matches make interpolation easier. This way, our feature descriptor supports the whole pipeline. Again, we did not change anything but the descriptor, not even the distance function that is used to compare the feature descriptors. CPM \cite{hu2016efficient} for example uses the sum of absolute difference as feature distance, while our network was trained using the L2 distance. 
Overall, our SDC features reduce the outliers during optical flow matching by up to 50~\%.

\paragraph{Matching-based Scene Flow Algorithms.} \label{sec:application:sceneflow}
SceneFlowFields (SFF) \cite{schuster2018sceneflowfields} is the stereo extension of FlowFields \cite{bailer2015flow,schuster2018combining} to estimate 3D motion. The pipeline is comparable to FF++ except for one additional refinement step that is used in SFF where the authors estimate the ego-motion to adjust the sceneflow of the static scene. We evaluate all intermediate results and present them in \cref{tab:sff}.

Similar to our experiments on stereo and optical flow, our SDC features improve scene flow matching significantly which results in almost half the percentage of outliers and endpoint errors. This effect can be maintained throughout the whole pipeline for almost all image regions. As before, outlier filtering at the foreground regions (\textit{fg}) of KITTI seems to be more difficult with SDC features which could probably be solved by adjusting the consistency threshold. The minor decrease in correctness of the filtered matches might again be acceptable when considering that SDC features increase the filtered density from 43.6~\% to 67.0~\% and from 36.4~\% to 56.0~\% in non-occluded (\textit{noc}) and \textit{all} image regions (cf. \cref{fig:visual_comparison}). Our SDC features improve scene flow matching over all image regions (including un-matchable, occluded areas) by more than 10 percent points which corresponds to 25~\% less outliers after matching.

\section{Conclusion}  \label{sec:conclusion}
Based on the observation that dilated convolution is related to sub-scale filtering, we have designed a novel layer by stacking multiple parallel dilated convolutions (SDC). These SDC layers have been combined to a new architecture that can be used for image feature description. For all experiments, we have used only a single unified network for all data sets and algorithms. Our SDC features have outperformed heuristic image descriptors like SIFT and other descriptor networks from previous works in terms of accuracy and robustness. In a second set of experiments, we have applied our SDC feature network for different matching tasks on many diverse data sets and have shown that our deep descriptor improves matching for stereo, optical flow, and scene flow drastically yielding a better final result in the majority of cases.

\section*{Acknowledgments}
This work was partially funded by the BMW Group and partly by the Federal Ministry of Education and Research Germany in the project VIDETE (01IW18002).

{\small
\bibliographystyle{ieee_fullname}
\bibliography{bib}
}

\end{document}

%%%%%%%%% TITLE
\title{Supplementary Material\\{\small SDC -- Stacked Dilated Convolution: A Unified Descriptor Network for Dense Matching Tasks}}

\author{René Schuster\textsuperscript{1} \hspace{0.5cm} Oliver Wasenmüller\textsuperscript{1} \hspace{0.5cm} Christian Unger\textsuperscript{2} \hspace{0.5cm} Didier Stricker\textsuperscript{1}\\
\textsuperscript{1}DFKI - German Research Center for Artificial Intelligence \hspace{0.5cm} \textsuperscript{2}BMW Group \\
{\tt\small firstname.lastname@\string{bmw,dfki\string}.de}
% \and
% Christian Unger\\
% BMW Group\\
% {\tt\small christian.unger@bmw.de}
}

\maketitle
\ifcvprfinal\thispagestyle{empty}\fi

%%%%%%%%% BODY TEXT
\section*{Introduction}
This is the supplementary material to our paper \textit{SDC -- Stacked Dilated Convolution: A Unified Descriptor Network for Dense Matching Tasks}. It covers the following contents that are mentioned in the main paper:
\begin{itemize}[noitemsep,topsep=1pt,label={\tiny\raisebox{0.75ex}{\textbullet}}]
	\item Comparison between multiple, parallel, dilated convolutions and a single convolution with a larger sparse kernel.
	\item Comparison of different hyper-parameters for our novel SDC network architecture.
	\item Detailed description for our sampling of training data with visual examples for sampled patch triplets.
	\item More visual results of SDC feature matching with different algorithms on different data sets.
\end{itemize}

\section{Large Sparse Convolution}
\begin{figure}[h]
	\centering
	\includegraphics[width=0.95\columnwidth]{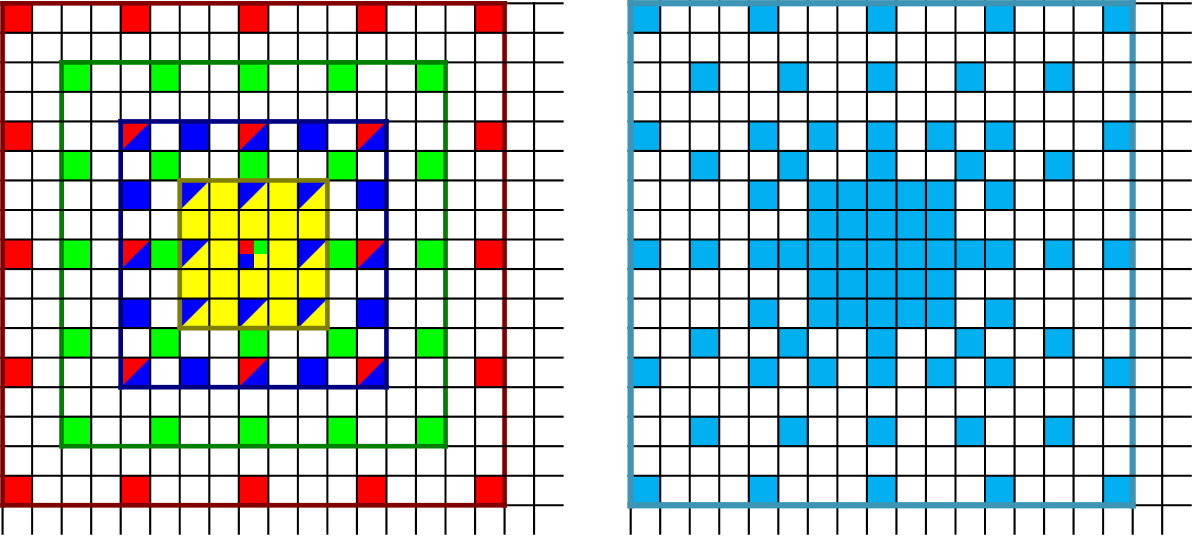}
	\hspace*{\fill} SDC Design \hfill \hfill Single Sparse Kernel \hspace*{\fill}
	\caption{Two variants of sparse $17 \times 17$ kernels. 4 parallel dilated $5 \times 5$ kernels (left), and a single kernel (right).}
	\label{fig:sdc_vs_large_sparse}
\end{figure}

Multiple dilated convolutions in parallel are similar to a single larger convolution with a sparse kernel (see \cref{fig:sdc_vs_large_sparse}). The difference is that pixels where the parallel kernels overlap are only considered once in a single kernel (for $3 \times 3$ kernels this will be the center pixel only) and that a single kernel will merge all information into a single output. With our parallel convolutions we have the choice to add them, stack them, or combine them as we want.

\begin{figure}[h]
	\centering
	\includegraphics[width=1\columnwidth]{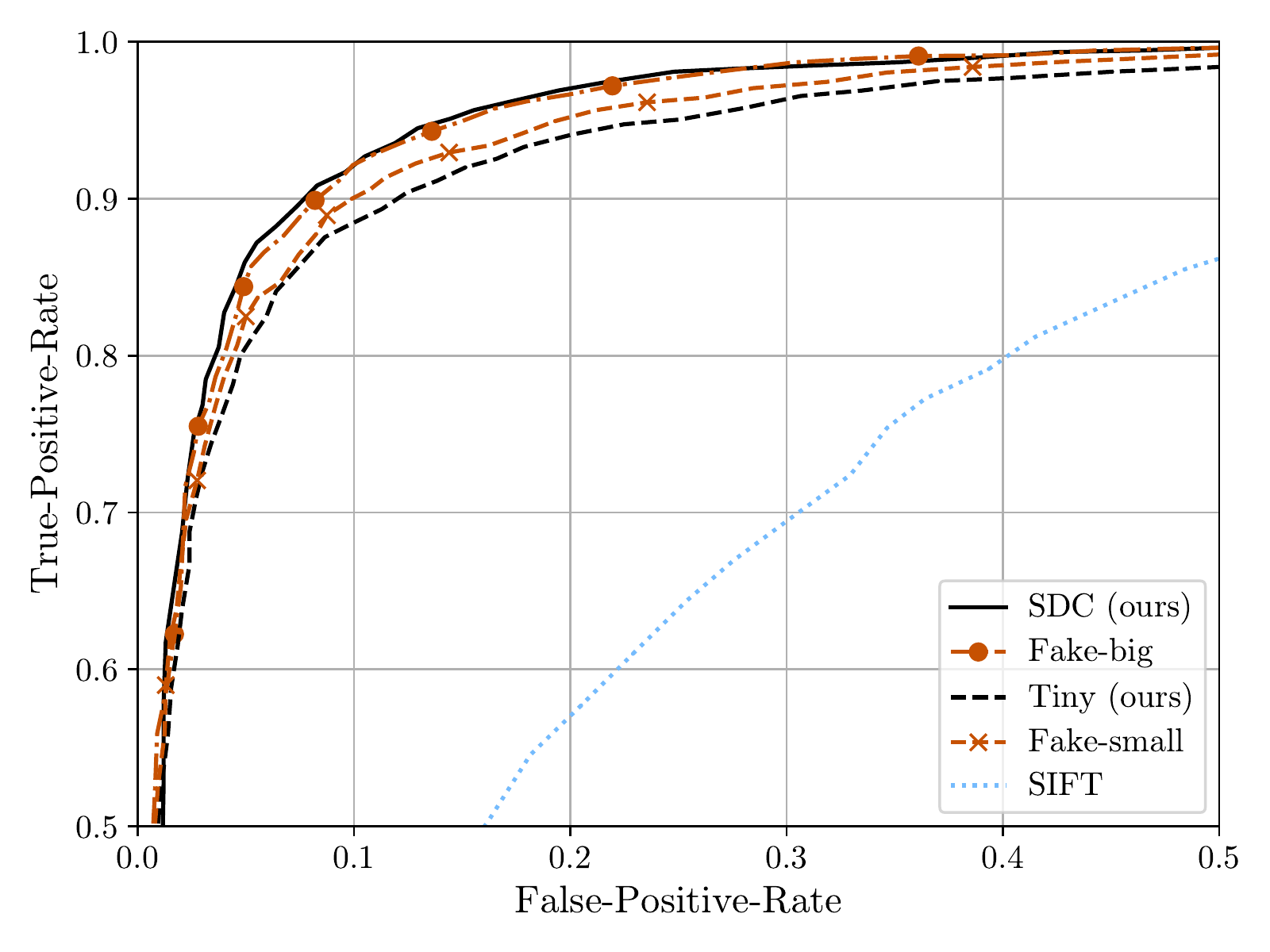}
	\caption{ROC curves for our SDC networks and the two \textit{Fake} variants.}
	\label{fig:roc_supp}
\end{figure}

\begin{figure}[h]
	\centering
	\includegraphics[width=1\columnwidth]{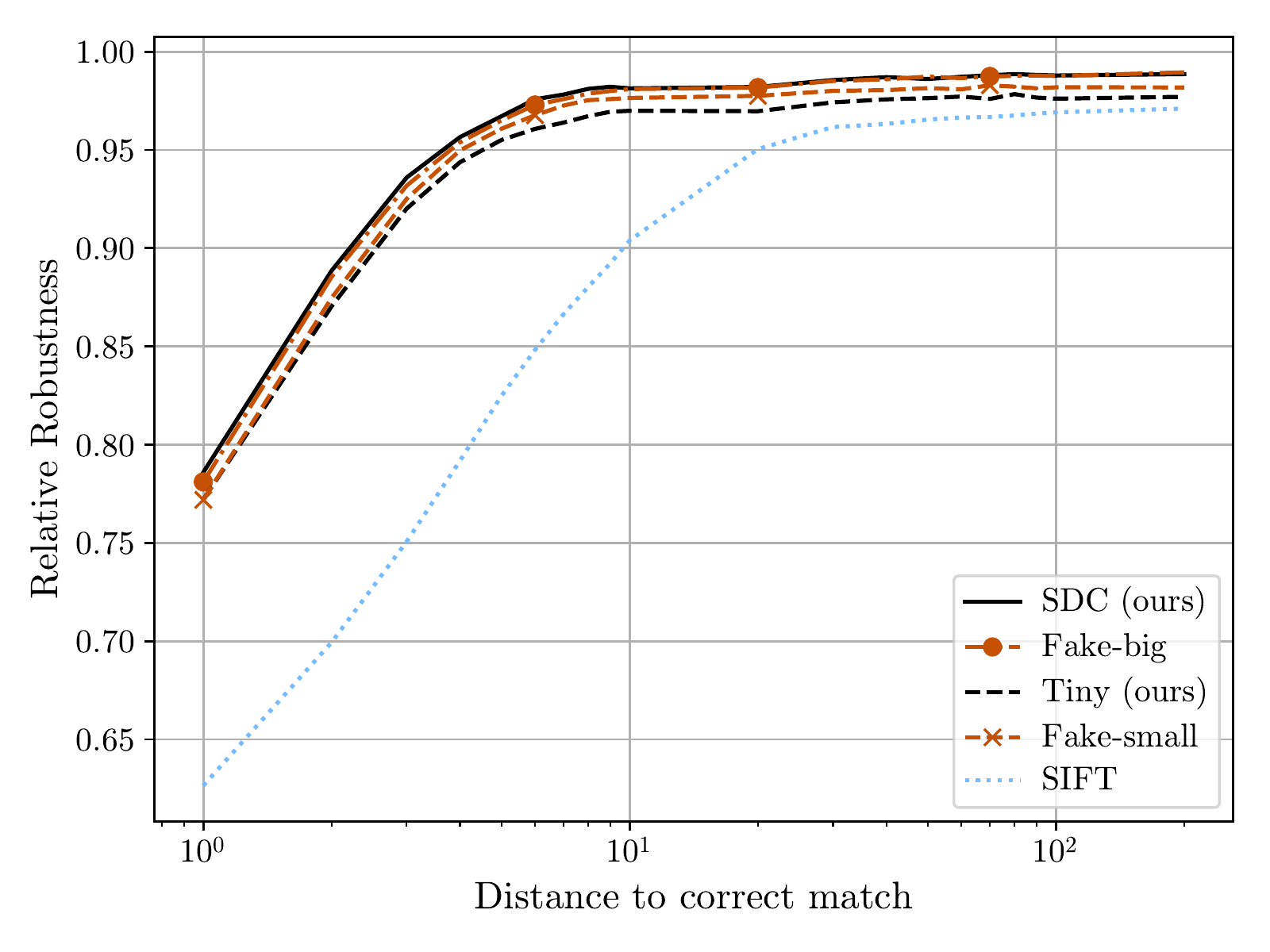}
	\caption{Robustness for our SDC networks and the two \textit{Fake} variants.}
	\label{fig:robustness_supp}
\end{figure}

For comparison of both approaches, we design two variants which use single, larger convolutions with the exact same receptive field as our SDC network. The first one produces the same output dimensions at each layer, \ie $64, 64, 128, 256, \text{and } 128$ feature channels. This results in a 4 times larger network compared to our SDC network approximately (disregarding overlapping pixel positions). The second variant is designed to achieve the same network size as our SDC network which results in 4 times less output channels per layer, \ie $16, 16, 32, 64, \text{and } 32$. We call these networks \textit{Fake-big} and \textit{Fake-small} respectively, because these networks try to imitate our original SDC design. The complete architectures look like this:
\begin{itemize}[noitemsep,topsep=1pt,label={\tiny\raisebox{0.75ex}{\textbullet}},leftmargin=*]
	\item \textit{Fake-big}: SparseConv(17,64,1,1)--SparseConv(17,64,1,1)--SparseConv(17,128,1,1)--SparseConv(17,256,1,1)--SparseConv(17,128,1,1)
	\item \textit{Fake-small}: SparseConv(17,16,1,1)--SparseConv(17,16,1,1)--SparseConv(17,32,1,1)--SparseConv(17,64,1,1)--SparseConv(17,32,1,1)
\end{itemize}
Sparsity is enforced according to the pattern shown in \cref{fig:sdc_vs_large_sparse}. 
The numbers parameterizing the convolutions are explained in Section 4.1 of the main paper.
Accuracy with network characteristics, ROC curves, and Robustness are evaluated in \cref{tab:accuracy_supp,fig:roc_supp,fig:robustness_supp}. These three metrics are explained in Section 4.1 of the main paper.

\begin{table}[h]
	\centering
	\caption{Comparison of the accuracy for our SDC networks and the \textit{Fake} variants that use single, sparse convolutions along with receptive field size (RF), number of parameters (Size) and accumulated sub-sampling factor due to striding.} \label{tab:accuracy_supp}
	\resizebox{\columnwidth}{!}{\begin{tabular}{c || c || c | c | c }
		\textbf{Network} & \textbf{Accuracy} & \textbf{RF} & \textbf{Size} & \textbf{Factor} \Bstrut\\
		\hline
		SDC (\textbf{Ours}) & \textbf{97.2 \%} & 81 & 1.95 M & 1\Tstrut\\
		Fake-big & 97.0 \% & 81 & 6.3 M & 1 \\
		Tiny (\textbf{Ours}) & 96.0 \% & 25 & \textbf{0.12 M} & 1 \\
		Fake-small & 96.0 \% & 81 & 0.4 M & 1 \\
	\end{tabular}}
\end{table}

In all evaluations, our SDC network outperforms the \textit{Fake} networks. Considering that \textit{Fake-big} is a much bigger network, this is even more evidence that our design is very powerful. The concatenation of multi-scale features and the mixture of multi-scale information at every level is beneficial for image description.

\begin{figure}[t]
	\centering
	\includegraphics[width=1\columnwidth]{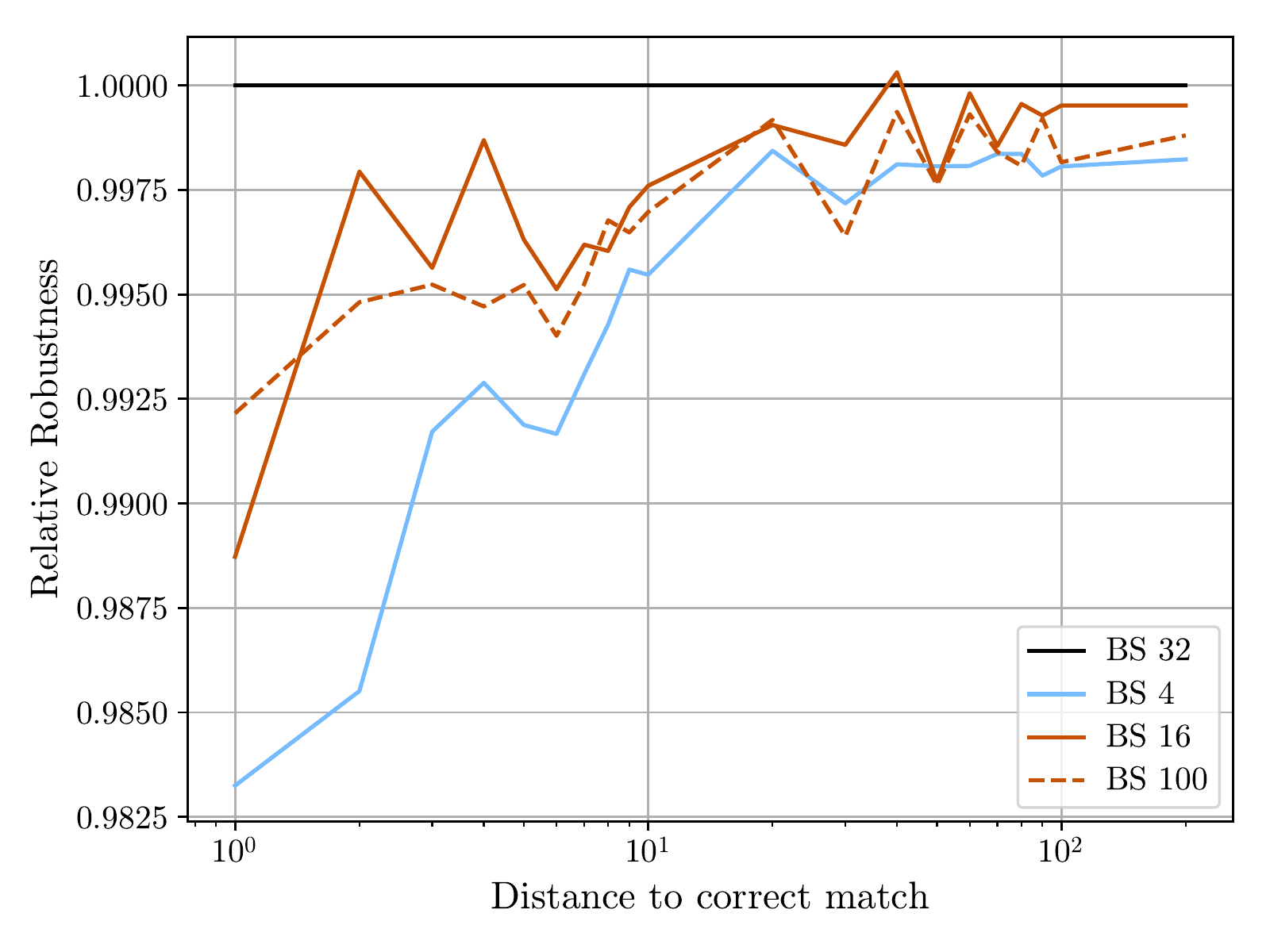}
	\caption{Relative robustness for our SDC network trained with different batch sizes (BS).}
	\label{fig:batchsize}
\end{figure}

\begin{figure}[t]
	\centering
	\includegraphics[width=1\columnwidth]{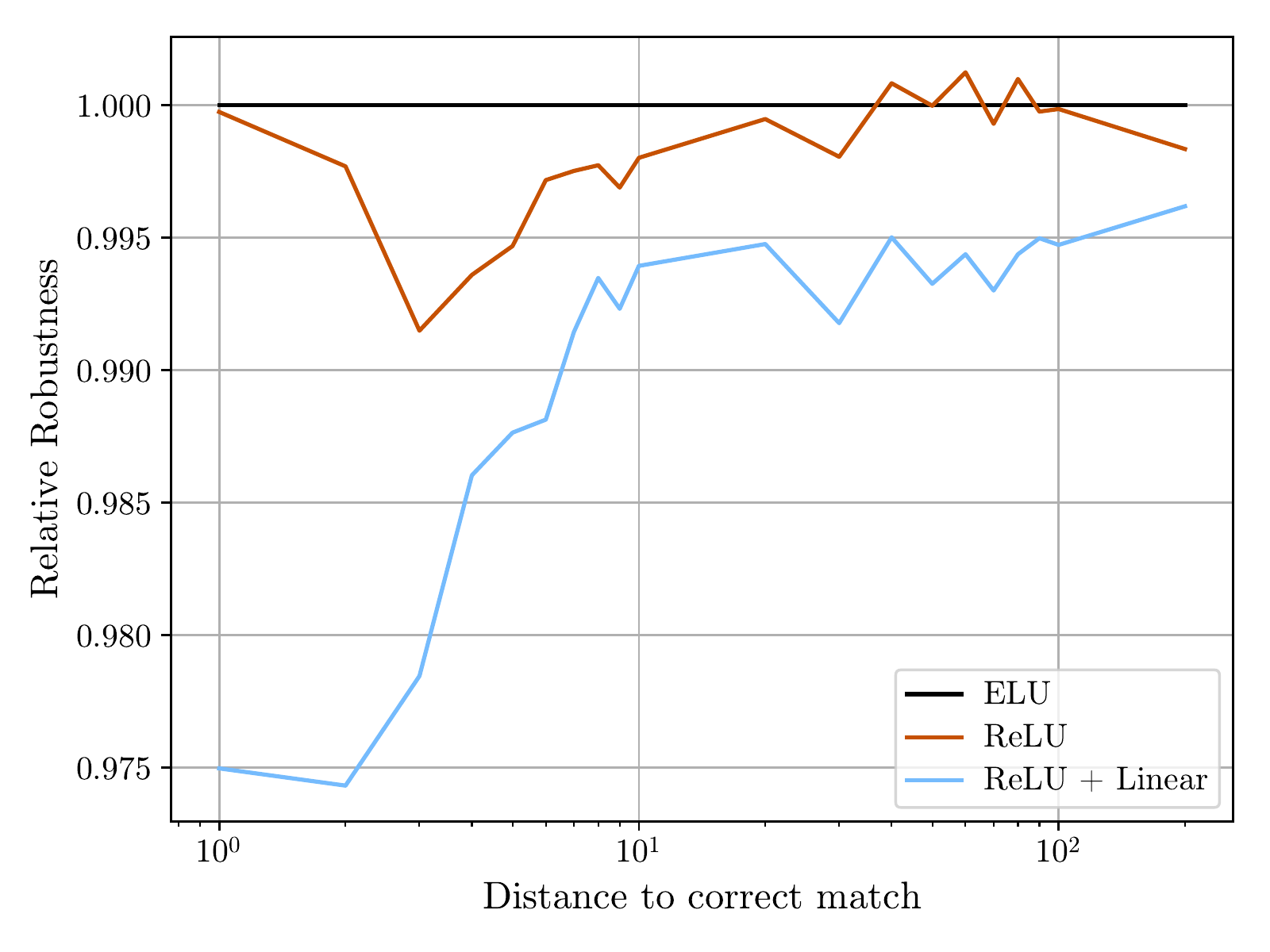}
	\caption{Relative robustness for our SDC network with different activation functions.}
	\label{fig:activation}
\end{figure}

\begin{figure}[t]
	\centering
	\includegraphics[width=1\columnwidth]{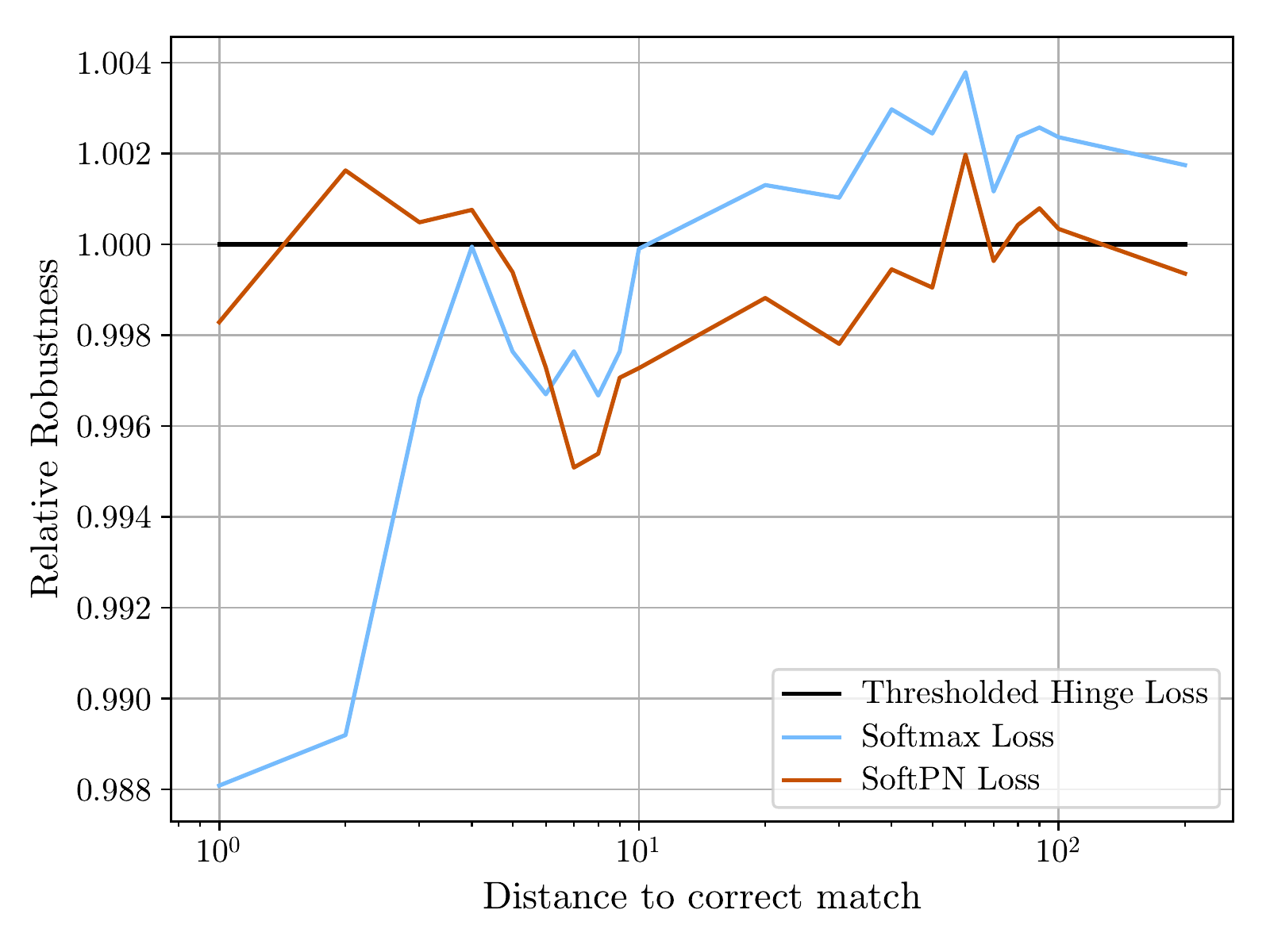}
	\caption{Relative robustness for our SDC network trained with different triplet loss functions.}
	\label{fig:loss}
\end{figure}

\section{Design Decisions}
Variation of depth and width of our SDC network structure is covered by our \textit{Tiny} version. In the following, we will vary batch size, activation functions, and loss function and compare them. We will alter one parameter at a time and compare the robustness to our original design. For better comparison, we introduce relative robustness which is the robustness ratio of a model and a reference model. A relative robustness greater than $1$ means that the model is better than the reference model. We will use our original SDC network as reference which will result in a baseline of 100 \% of relative robustness for this model.

Results for different batch sizes are given in \cref{fig:batchsize}. The impact of the batch size is minor. Even for very small mini batches, the relative robustness drops by less than 2 percent points. Increasing the batch size is not improving the performance either.

For alternative activation functions, we consider ReLU \cite{nair2010rectified} instead of ELU \cite{clevert2015fast}. Since our final feature vectors are normalized to unit range, the rectification of ReLU restricts the feature space to the non-negative orthant of the 128-dimensional hypercube which is only $2^{-128}$ of the full volume. Therefore, we also train and compare a network with ReLU and linear activation in the last layer. The comparison is shown in \cref{fig:activation}.

As mentioned in the main paper where we use triplet training with a thresholded hinge embedding loss \cite{bailer2017cnn}, we have also experimented with the softmax loss of \cite{hoffer2015deep} and the PN loss of \cite{balntas2016pn} (see \cref{fig:loss}). Both alternative losses perform better for some displacements, and worse for others. However, the difference in robustness is not significant and training with these real triplet losses is much less stable forcing us to use a lower learning rate and thus increasing overall training time.

\section{Training Data}
\begin{figure}[h]
	\centering
	\includegraphics[width=0.95\columnwidth]{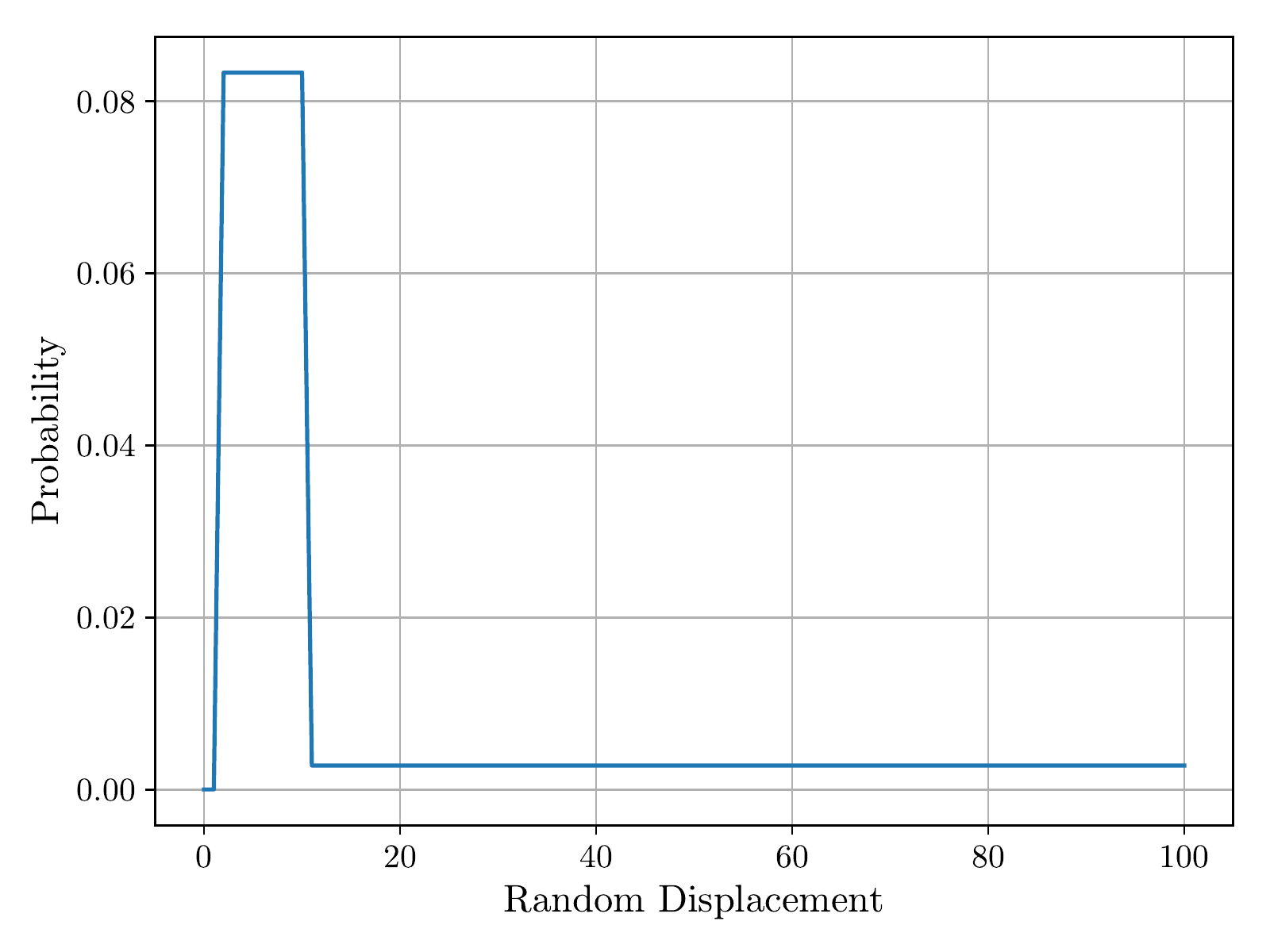}
	\caption{Probability distribution of the random offset used to generate the negative patch correspondence.}
	\label{fig:offset}
\end{figure}

To train a single unified descriptor network that can be used for any algorithm on any domain, we use multiple data sets with very diverse characteristics. They differ in number of sequences and sequence lengths, image resolution, available ground truth, ground truth masks, and color channels. We have tried to consider all these factors to avoid imbalance and over-representations in our training data. In the end, we sample reference images with the following probabilities for each data set:
\begin{itemize}[noitemsep,topsep=1pt,label={\tiny\raisebox{0.75ex}{\textbullet}}]
	\item KITTI \cite{menze2015object}: 0.5
	\item Sintel \cite{butler2012sintel}: 0.175
	\item Middlebury Optical Flow\cite{baker2011database}: 0.025
	\item Middlebury Stereo \cite{scharstein2002taxonomy}: 0.05
	\item HD1K \cite{kondermann2016hci}: 0.175
	\item ETH3D \cite{schops2017multi}: 0,075
\end{itemize}
We do not sample the same image twice until all images of the respective data set were selected, \ie within each data set, we use all images equally often.
For each image in each epoch, 100 randomly sampled reference patches are selected. If a data set provides ground truth for multiple tasks (\eg KITTI \cite{menze2015object}), we randomly select one of the tasks from stereo, optical flow, or scene flow. The reference patches are sampled from pixel positions where ground truth exists and where the ground truth displacement points to a visible position in the corresponding view, \ie occlusions and out-of-bound displacements are excluded wherever possible. For each of the 100 reference patches, we extract the corresponding matching patch according to the ground truth. We pad the image with reflection at image boundaries and use bilinear interpolation at sub-pixel positions.
The negative patch is extracted by altering the ground truth displacement with a random offset. This random offset is at least 2 pixels and at most 100 pixels large in magnitude. For stereo correspondences, the displacement and the random offset are 1-dimensional along the horizontal direction according to the epipolar constraint. Other correspondences have 2-dimensional displacement, \ie circular around the ground truth correspondence.
Since close-by correspondences are harder to distinguish, we sample the random offset non-uniformly. In detail, we split the random offset into two ranges, a close one ($\left[2,10\right]$ pixels) and a distant range ($\left]10,100\right]$ pixels). The close range is selected 3 times more often than the far range and within each range, we sample uniformly. This leads to the overall probability distribution for our random negative displacements shown in \cref{fig:offset}.

Each image is processed completely (all 100 patch triplets) before selecting the next image pair from one of the data sets to reduce IO operations in our training data pipeline. We then shuffle chunks of 3200 triplets to reintroduce randomness across data sets and images.
As mentioned in the main paper, the input patches are normalized according to the mean pixel and variance of the complete training data. The mean pixel is [0.3534, 0.3448, 0.3295] and the mean standard deviation is [0.2492, 0.2465, 0.2446] for the red, green, and blue color channels respectively.

\begin{figure*}[p]
	\centering
	\begin{subfigure}[c]{0.48\textwidth}
		\centering
		\begin{subfigure}[c]{0.3\textwidth}
			\centering Positive\\Patch
		\end{subfigure}
		\begin{subfigure}[c]{0.3\textwidth}
			\centering Reference\\Patch
		\end{subfigure}
		\begin{subfigure}[c]{0.3\textwidth}
			\centering Negative\\Patch
		\end{subfigure}\\%
		\vspace{1mm}%
		\includegraphics[width=0.3\textwidth]{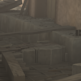}
		\includegraphics[width=0.3\textwidth]{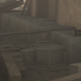}
		\includegraphics[width=0.3\textwidth]{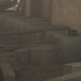}
	\end{subfigure}
	\hfill
	\begin{subfigure}[c]{0.48\textwidth}
		\centering
		\begin{subfigure}[c]{0.3\textwidth}
			\centering Positive\\Patch
		\end{subfigure}
		\begin{subfigure}[c]{0.3\textwidth}
			\centering Reference\\Patch
		\end{subfigure}
		\begin{subfigure}[c]{0.3\textwidth}
			\centering Negative\\Patch
		\end{subfigure}\\%
		\vspace{1mm}%
		\includegraphics[width=0.3\textwidth]{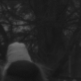}
		\includegraphics[width=0.3\textwidth]{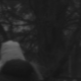}
		\includegraphics[width=0.3\textwidth]{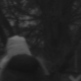}
	\end{subfigure}\\%
	\vspace{2mm}%
	\begin{subfigure}[c]{0.48\textwidth}
		\centering
		\includegraphics[width=0.3\textwidth]{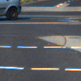}
		\includegraphics[width=0.3\textwidth]{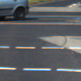}
		\includegraphics[width=0.3\textwidth]{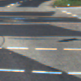}
	\end{subfigure}
	\hfill
	\begin{subfigure}[c]{0.48\textwidth}
		\centering
		\includegraphics[width=0.3\textwidth]{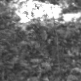}
		\includegraphics[width=0.3\textwidth]{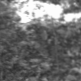}
		\includegraphics[width=0.3\textwidth]{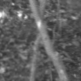}
	\end{subfigure}\\%
	\vspace{2mm}%
	\begin{subfigure}[c]{0.48\textwidth}
		\centering
		\includegraphics[width=0.3\textwidth]{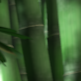}
		\includegraphics[width=0.3\textwidth]{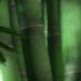}
		\includegraphics[width=0.3\textwidth]{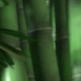}
	\end{subfigure}
	\hfill
	\begin{subfigure}[c]{0.48\textwidth}
		\centering
		\includegraphics[width=0.3\textwidth]{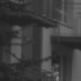}
		\includegraphics[width=0.3\textwidth]{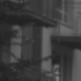}
		\includegraphics[width=0.3\textwidth]{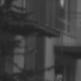}
	\end{subfigure}\\%
	\vspace{2mm}%
	\begin{subfigure}[c]{0.48\textwidth}
		\centering
		\includegraphics[width=0.3\textwidth]{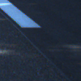}
		\includegraphics[width=0.3\textwidth]{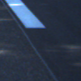}
		\includegraphics[width=0.3\textwidth]{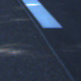}
	\end{subfigure}
	\hfill
	\begin{subfigure}[c]{0.48\textwidth}
		\centering
		\includegraphics[width=0.3\textwidth]{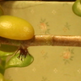}
		\includegraphics[width=0.3\textwidth]{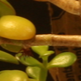}
		\includegraphics[width=0.3\textwidth]{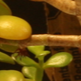}
	\end{subfigure}\\%
	\vspace{2mm}%
	\begin{subfigure}[c]{0.48\textwidth}
		\centering
		\includegraphics[width=0.3\textwidth]{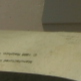}
		\includegraphics[width=0.3\textwidth]{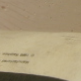}
		\includegraphics[width=0.3\textwidth]{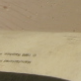}
	\end{subfigure}
	\hfill
	\begin{subfigure}[c]{0.48\textwidth}
		\centering
		\includegraphics[width=0.3\textwidth]{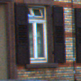}
		\includegraphics[width=0.3\textwidth]{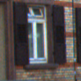}
		\includegraphics[width=0.3\textwidth]{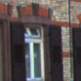}
	\end{subfigure}\\%
	\vspace{2mm}%
	\begin{subfigure}[c]{0.48\textwidth}
		\centering
		\includegraphics[width=0.3\textwidth]{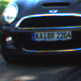}
		\includegraphics[width=0.3\textwidth]{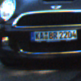}
		\includegraphics[width=0.3\textwidth]{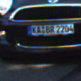}
	\end{subfigure}
	\hfill
	\begin{subfigure}[c]{0.48\textwidth}
		\centering
		\includegraphics[width=0.3\textwidth]{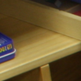}
		\includegraphics[width=0.3\textwidth]{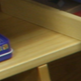}
		\includegraphics[width=0.3\textwidth]{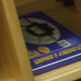}
	\end{subfigure}\\%
	\vspace{2mm}%
	\begin{subfigure}[c]{0.48\textwidth}
		\centering
		\includegraphics[width=0.3\textwidth]{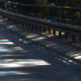}
		\includegraphics[width=0.3\textwidth]{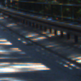}
		\includegraphics[width=0.3\textwidth]{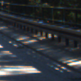}
	\end{subfigure}
	\hfill
	\begin{subfigure}[c]{0.48\textwidth}
		\centering
		\includegraphics[width=0.3\textwidth]{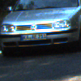}
		\includegraphics[width=0.3\textwidth]{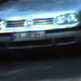}
		\includegraphics[width=0.3\textwidth]{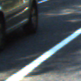}
	\end{subfigure}
	\caption{Randomly sampled training triplets. For each reference patch, we select a corresponding patch according to the ground truth displacement and a negative correspondence by adding a random offset to the ground truth.}
	\label{fig:patches}
\end{figure*}

Smaller displacements (less than 2 pixels) are not considered for several reasons. Minimal changes in appearance might confuse the network, rounding and interpolation introduce small inaccuracies, and most applications tolerate a matching accuracy less than 2 pixels endpoint error.
Visual examples of our training triplets are given in \cref{fig:patches}.

The 200 training images from KITTI \cite{menze2015object}, are randomly split into a subset for actual training (70~\%), one for validation (20~\%), and one for testing in our experiments section (10~\%). This is the exact list of sequences for each subset:
\begin{itemize}[noitemsep,topsep=1pt,label={\tiny\raisebox{0.75ex}{\textbullet}}]
	\item Training:  0,   1,   3,   5,   6,   7,   8,   9,  10,  11,  12,  13,  14,
        15,  18,  19,  20,  21,  22,  24,  25,  27,  28,  29,  30,  31,
        33,  34,  35,  37,  38,  39,  40,  41,  43,  44,  45,  47,  49,
        50,  51,  54,  55,  56,  58,  59,  62,  63,  66,  67,  68,  70,
        71,  74,  75,  76,  78,  79,  82,  83,  85,  86,  87,  88,  89,
        90,  93,  95,  96,  97,  99, 101, 102, 103, 104, 105, 107, 109,
       111, 113, 114, 116, 117, 118, 120, 123, 125, 128, 129, 130, 132,
       134, 135, 137, 138, 139, 140, 141, 143, 144, 145, 147, 148, 149,
       150, 151, 152, 154, 155, 156, 157, 160, 161, 162, 163, 164, 165,
       167, 168, 169, 170, 171, 172, 175, 177, 178, 179, 180, 182, 183,
       185, 187, 188, 189, 191, 194, 195, 197, 198, 199
	\item Validation: 2,  16,  17,  23,  26,  32,  36,  48,  52,  53,  57,  60,  61,
        64,  69,  72,  73,  77,  80,  81,  84,  91, 100, 108, 110, 112,
       122, 126, 127, 131, 133, 136, 142, 153, 158, 159, 166, 176, 192,
       196
	\item Testing: 4,  42,  46,  65,  92,  94,  98, 106, 115, 119, 121, 124, 146,
       173, 174, 181, 184, 186, 190, 193
\end{itemize}

\section{Visual Results}
\begin{figure}[h]
	\centering
	\includegraphics[width=0.95\columnwidth]{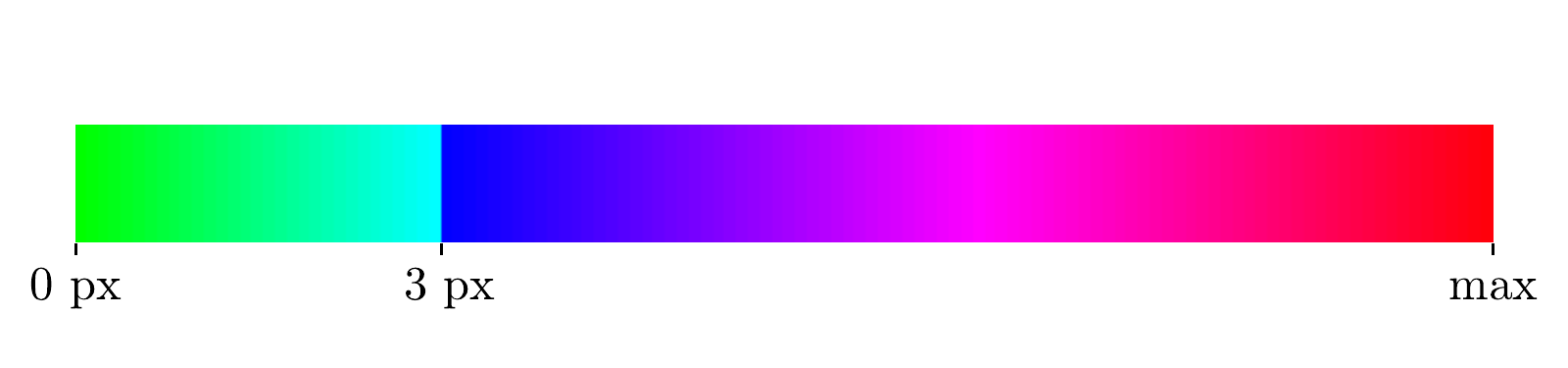}
	\caption{Color encoding of the endpoint error in the error maps in \crefrange{fig:visual_results:elas_kitti}{fig:visual_results:sff_kitti}.}
	\label{fig:errorcode}
\end{figure}

In addition to the quantitative results for ELAS \cite{geiger2010elas}, SGM \cite{hirschmuller2008SGM}, CPM \cite{hu2016efficient}, FlowFields++ \cite{schuster2018ffpp}, and SceneFlowFields \cite{schuster2018sceneflowfields} presented in the main paper in Tables 2 to 5, we provide visual examples on the different data sets in \cref{fig:visual_results}. For each combination of matching algorithm and data set, we show the corresponding images, the estimated matches using the original feature descriptor, and matches computed with our SDC features. Please note that we do not tune any algorithm for our new descriptor. We replace the feature descriptor and change nothing else. For each matching result, we also give an error map indicating where the estimate exceeds an endpoint error of 3 pixels (\textgreater3px). The color encoding for these error maps is shown in \cref{fig:errorcode}.

In case of scene flow (\cref{fig:visual_results:sff_kitti}), we split the visualization of the estimated result into optical flow and two disparity maps.
Since CPM \cite{hu2016efficient} computes sparse matches in $3 \times 3$ blocks only, we dilate the visualization for matches and error maps.
More results can be found in our supplementary video \url{https://youtu.be/RoxfVdfqWpY}.

\begin{figure*}[p]
	\centering
	\begin{subfigure}[c]{1.0\textwidth}
		\centering
		\begin{subfigure}[c]{0.45\textwidth}
			\centering
			\includegraphics[width=1\textwidth]{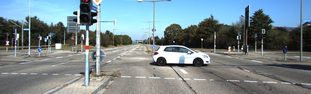}
		\end{subfigure}
		\begin{subfigure}[c]{0.45\textwidth}
			\centering
			\includegraphics[width=1\textwidth]{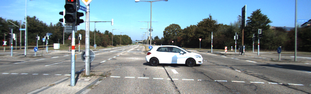}
		\end{subfigure}\\%
		\vspace{0.5mm}%
		\begin{subfigure}[c]{0.45\textwidth}
			\centering \small Left Image
		\end{subfigure}
		\begin{subfigure}[c]{0.45\textwidth}
			\centering \small Right Image
		\end{subfigure}\\%
		\vspace{1.5mm}%
		\begin{subfigure}[c]{0.45\textwidth}
			\centering
			\includegraphics[width=1\textwidth]{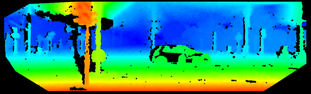}
		\end{subfigure}
		\begin{subfigure}[c]{0.45\textwidth}
			\centering
			\includegraphics[width=1\textwidth]{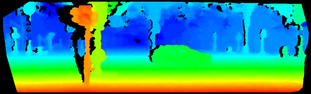}
		\end{subfigure}\\%
		\vspace{0.5mm}%
		\begin{subfigure}[c]{0.45\textwidth}
			\centering \small Original Result
		\end{subfigure}
		\begin{subfigure}[c]{0.45\textwidth}
			\centering \small Result with SDC
		\end{subfigure}\\%
		\vspace{1.5mm}%
		\begin{subfigure}[c]{0.45\textwidth}
			\centering
			\includegraphics[width=1\textwidth]{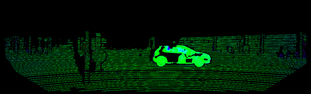}
		\end{subfigure}
		\begin{subfigure}[c]{0.45\textwidth}
			\centering
			\includegraphics[width=1\textwidth]{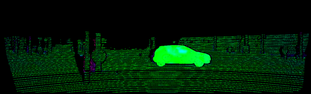}
		\end{subfigure}\\%
		\vspace{0.5mm}%
		\begin{subfigure}[c]{0.45\textwidth}
			\centering \small Error Map for Original Result
		\end{subfigure}
		\begin{subfigure}[c]{0.45\textwidth}
			\centering \small Error Map for Result with SDC
		\end{subfigure}		
		\caption{Stereo matching result for ELAS \cite{geiger2010elas} on KITTI \cite{menze2015object}. Quantitative results are given in Table 2 of the main paper.}
		\label{fig:visual_results:elas_kitti}
	\end{subfigure}\\%
	\vspace{10mm}%
	\begin{subfigure}[c]{1.0\textwidth}
		\centering
		\begin{subfigure}[c]{0.45\textwidth}
			\centering
			\includegraphics[width=1\textwidth]{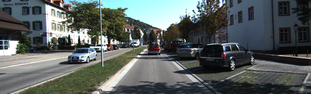}
		\end{subfigure}
		\begin{subfigure}[c]{0.45\textwidth}
			\centering
			\includegraphics[width=1\textwidth]{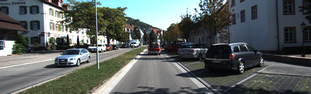}
		\end{subfigure}\\%
		\vspace{0.5mm}%
		\begin{subfigure}[c]{0.45\textwidth}
			\centering \small Left Image
		\end{subfigure}
		\begin{subfigure}[c]{0.45\textwidth}
			\centering \small Right Image
		\end{subfigure}\\%
		\vspace{1.5mm}%
		\begin{subfigure}[c]{0.45\textwidth}
			\centering
			\includegraphics[width=1\textwidth]{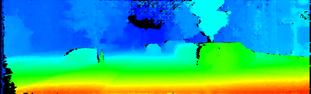}
		\end{subfigure}
		\begin{subfigure}[c]{0.45\textwidth}
			\centering
			\includegraphics[width=1\textwidth]{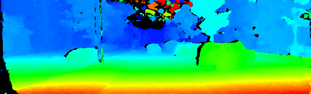}
		\end{subfigure}\\%
		\vspace{0.5mm}%
		\begin{subfigure}[c]{0.45\textwidth}
			\centering \small Original Result
		\end{subfigure}
		\begin{subfigure}[c]{0.45\textwidth}
			\centering \small Result with SDC
		\end{subfigure}\\%
		\vspace{1.5mm}%
		\begin{subfigure}[c]{0.45\textwidth}
			\centering
			\includegraphics[width=1\textwidth]{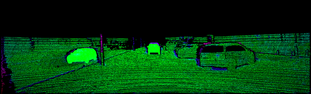}
		\end{subfigure}
		\begin{subfigure}[c]{0.45\textwidth}
			\centering
			\includegraphics[width=1\textwidth]{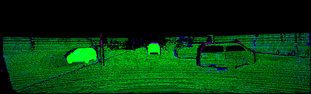}
		\end{subfigure}\\%
		\vspace{0.5mm}%
		\begin{subfigure}[c]{0.45\textwidth}
			\centering \small Error Map for Original Result
		\end{subfigure}
		\begin{subfigure}[c]{0.45\textwidth}
			\centering \small Error Map for Result with SDC
		\end{subfigure}		
		\caption{Stereo matching result for SGM \cite{hirschmuller2008SGM} on KITTI \cite{menze2015object}.  Quantitative results are given in Table 2 of the main paper.}
		\label{fig:visual_results:sgm_kitti}
	\end{subfigure}
\end{figure*}

\begin{figure*}[p]
	\ContinuedFloat
	\centering
	\begin{subfigure}[c]{1.0\textwidth}
		\centering
		\begin{subfigure}[c]{0.45\textwidth}
			\centering
			\includegraphics[width=1\textwidth]{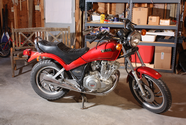}
		\end{subfigure}
		\begin{subfigure}[c]{0.45\textwidth}
			\centering
			\includegraphics[width=1\textwidth]{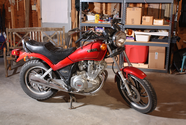}
		\end{subfigure}\\%
		\vspace{0.5mm}%
		\begin{subfigure}[c]{0.45\textwidth}
			\centering \small Left Image
		\end{subfigure}
		\begin{subfigure}[c]{0.45\textwidth}
			\centering \small Right Image
		\end{subfigure}\\%
		\vspace{1.5mm}%
		\begin{subfigure}[c]{0.45\textwidth}
			\centering
			\includegraphics[width=1\textwidth]{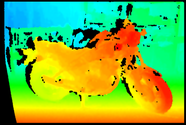}
		\end{subfigure}
		\begin{subfigure}[c]{0.45\textwidth}
			\centering
			\includegraphics[width=1\textwidth]{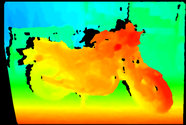}
		\end{subfigure}\\%
		\vspace{0.5mm}%
		\begin{subfigure}[c]{0.45\textwidth}
			\centering \small Original Result
		\end{subfigure}
		\begin{subfigure}[c]{0.45\textwidth}
			\centering \small Result with SDC
		\end{subfigure}\\%
		\vspace{1.5mm}%
		\begin{subfigure}[c]{0.45\textwidth}
			\centering
			\includegraphics[width=1\textwidth]{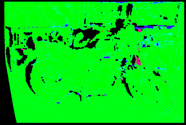}
		\end{subfigure}
		\begin{subfigure}[c]{0.45\textwidth}
			\centering
			\includegraphics[width=1\textwidth]{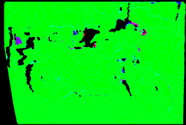}
		\end{subfigure}\\%
		\vspace{0.5mm}%
		\begin{subfigure}[c]{0.45\textwidth}
			\centering \small Error Map for Original Result
		\end{subfigure}
		\begin{subfigure}[c]{0.45\textwidth}
			\centering \small Error Map for Result with SDC
		\end{subfigure}		
		\caption{Stereo matching result for ELAS \cite{geiger2010elas} on Middlebury \cite{scharstein2002taxonomy}. Quantitative results are given in Table 2 of the main paper.}
		\label{fig:visual_results:elas_middlebury}
	\end{subfigure}
\end{figure*}

\begin{figure*}[p]
	\ContinuedFloat
	\centering
	\begin{subfigure}[c]{1.0\textwidth}
		\centering
		\begin{subfigure}[c]{0.45\textwidth}
			\centering
			\includegraphics[width=1\textwidth]{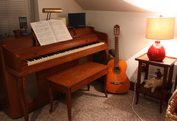}
		\end{subfigure}
		\begin{subfigure}[c]{0.45\textwidth}
			\centering
			\includegraphics[width=1\textwidth]{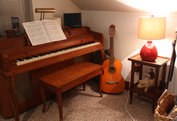}
		\end{subfigure}\\%
		\vspace{0.5mm}%
		\begin{subfigure}[c]{0.45\textwidth}
			\centering \small Left Image
		\end{subfigure}
		\begin{subfigure}[c]{0.45\textwidth}
			\centering \small Right Image
		\end{subfigure}\\%
		\vspace{1.5mm}%
		\begin{subfigure}[c]{0.45\textwidth}
			\centering
			\includegraphics[width=1\textwidth]{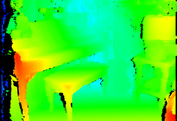}
		\end{subfigure}
		\begin{subfigure}[c]{0.45\textwidth}
			\centering
			\includegraphics[width=1\textwidth]{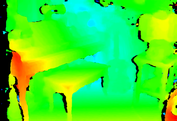}
		\end{subfigure}\\%
		\vspace{0.5mm}%
		\begin{subfigure}[c]{0.45\textwidth}
			\centering \small Original Result
		\end{subfigure}
		\begin{subfigure}[c]{0.45\textwidth}
			\centering \small Result with SDC
		\end{subfigure}\\%
		\vspace{1.5mm}%
		\begin{subfigure}[c]{0.45\textwidth}
			\centering
			\includegraphics[width=1\textwidth]{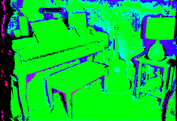}
		\end{subfigure}
		\begin{subfigure}[c]{0.45\textwidth}
			\centering
			\includegraphics[width=1\textwidth]{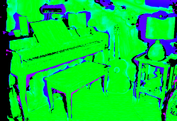}
		\end{subfigure}\\%
		\vspace{0.5mm}%
		\begin{subfigure}[c]{0.45\textwidth}
			\centering \small Error Map for Original Result
		\end{subfigure}
		\begin{subfigure}[c]{0.45\textwidth}
			\centering \small Error Map for Result with SDC
		\end{subfigure}		
		\caption{Stereo matching result for SGM \cite{hirschmuller2008SGM} on Middlebury \cite{scharstein2002taxonomy}. Quantitative results are given in Table 2 of the main paper.}
		\label{fig:visual_results:sgm_middlebury}
	\end{subfigure}
\end{figure*}

\begin{figure*}[p]
	\ContinuedFloat
	\centering
	\begin{subfigure}[c]{1.0\textwidth}
		\centering
		\begin{subfigure}[c]{0.45\textwidth}
			\centering
			\includegraphics[width=1\textwidth]{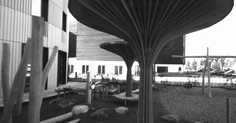}
		\end{subfigure}
		\begin{subfigure}[c]{0.45\textwidth}
			\centering
			\includegraphics[width=1\textwidth]{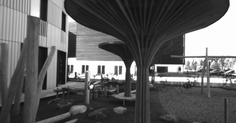}
		\end{subfigure}\\%
		\vspace{0.5mm}%
		\begin{subfigure}[c]{0.45\textwidth}
			\centering \small Left Image
		\end{subfigure}
		\begin{subfigure}[c]{0.45\textwidth}
			\centering \small Right Image
		\end{subfigure}\\%
		\vspace{1.5mm}%
		\begin{subfigure}[c]{0.45\textwidth}
			\centering
			\includegraphics[width=1\textwidth]{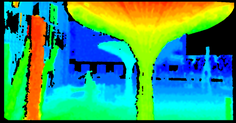}
		\end{subfigure}
		\begin{subfigure}[c]{0.45\textwidth}
			\centering
			\includegraphics[width=1\textwidth]{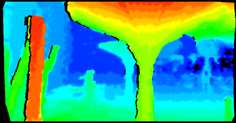}
		\end{subfigure}\\%
		\vspace{0.5mm}%
		\begin{subfigure}[c]{0.45\textwidth}
			\centering \small Original Result
		\end{subfigure}
		\begin{subfigure}[c]{0.45\textwidth}
			\centering \small Result with SDC
		\end{subfigure}\\%
		\vspace{1.5mm}%
		\begin{subfigure}[c]{0.45\textwidth}
			\centering
			\includegraphics[width=1\textwidth]{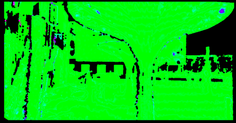}
		\end{subfigure}
		\begin{subfigure}[c]{0.45\textwidth}
			\centering
			\includegraphics[width=1\textwidth]{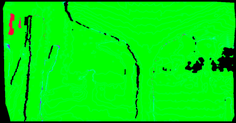}
		\end{subfigure}\\%
		\vspace{0.5mm}%
		\begin{subfigure}[c]{0.45\textwidth}
			\centering \small Error Map for Original Result
		\end{subfigure}
		\begin{subfigure}[c]{0.45\textwidth}
			\centering \small Error Map for Result with SDC
		\end{subfigure}		
		\caption{Stereo matching result for ELAS \cite{geiger2010elas} on ETH3D \cite{schops2017multi}. Quantitative results are given in Table 2 of the main paper.}
		\label{fig:visual_results:elas_eth3d}
	\end{subfigure}
\end{figure*}

\begin{figure*}[p]
	\ContinuedFloat
	\centering
	\begin{subfigure}[c]{1.0\textwidth}
		\centering
		\begin{subfigure}[c]{0.45\textwidth}
			\centering
			\includegraphics[width=1\textwidth]{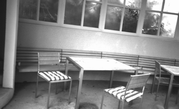}
		\end{subfigure}
		\begin{subfigure}[c]{0.45\textwidth}
			\centering
			\includegraphics[width=1\textwidth]{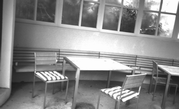}
		\end{subfigure}\\%
		\vspace{0.5mm}%
		\begin{subfigure}[c]{0.45\textwidth}
			\centering \small Left Image
		\end{subfigure}
		\begin{subfigure}[c]{0.45\textwidth}
			\centering \small Right Image
		\end{subfigure}\\%
		\vspace{1.5mm}%
		\begin{subfigure}[c]{0.45\textwidth}
			\centering
			\includegraphics[width=1\textwidth]{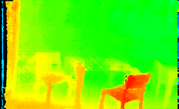}
		\end{subfigure}
		\begin{subfigure}[c]{0.45\textwidth}
			\centering
			\includegraphics[width=1\textwidth]{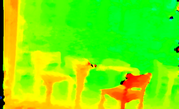}
		\end{subfigure}\\%
		\vspace{0.5mm}%
		\begin{subfigure}[c]{0.45\textwidth}
			\centering \small Original Result
		\end{subfigure}
		\begin{subfigure}[c]{0.45\textwidth}
			\centering \small Result with SDC
		\end{subfigure}\\%
		\vspace{1.5mm}%
		\begin{subfigure}[c]{0.45\textwidth}
			\centering
			\includegraphics[width=1\textwidth]{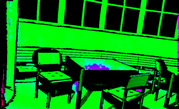}
		\end{subfigure}
		\begin{subfigure}[c]{0.45\textwidth}
			\centering
			\includegraphics[width=1\textwidth]{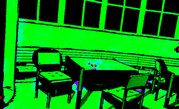}
		\end{subfigure}\\%
		\vspace{0.5mm}%
		\begin{subfigure}[c]{0.45\textwidth}
			\centering \small Error Map for Original Result
		\end{subfigure}
		\begin{subfigure}[c]{0.45\textwidth}
			\centering \small Error Map for Result with SDC
		\end{subfigure}		
		\caption{Stereo matching result for SGM \cite{hirschmuller2008SGM} on ETH3D \cite{schops2017multi}. Quantitative results are given in Table 2 of the main paper.}
		\label{fig:visual_results:sgm_eth3d}
	\end{subfigure}
\end{figure*}

\begin{figure*}[p]
	\ContinuedFloat
	\centering
	\begin{subfigure}[c]{1.0\textwidth}
		\centering
		\begin{subfigure}[c]{0.45\textwidth}
			\centering
			\includegraphics[width=1\textwidth]{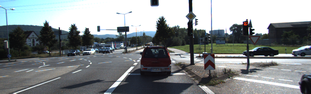}
		\end{subfigure}
		\begin{subfigure}[c]{0.45\textwidth}
			\centering
			\includegraphics[width=1\textwidth]{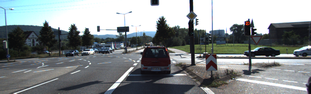}
		\end{subfigure}\\%
		\vspace{0.5mm}%
		\begin{subfigure}[c]{0.45\textwidth}
			\centering \small First Image
		\end{subfigure}
		\begin{subfigure}[c]{0.45\textwidth}
			\centering \small Second Image
		\end{subfigure}\\%
		\vspace{1.5mm}%
		\begin{subfigure}[c]{0.45\textwidth}
			\centering
			\includegraphics[width=1\textwidth]{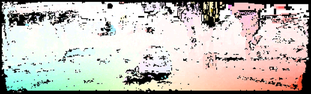}
		\end{subfigure}
		\begin{subfigure}[c]{0.45\textwidth}
			\centering
			\includegraphics[width=1\textwidth]{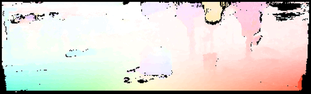}
		\end{subfigure}\\%
		\vspace{0.5mm}%
		\begin{subfigure}[c]{0.45\textwidth}
			\centering \small Original Result
		\end{subfigure}
		\begin{subfigure}[c]{0.45\textwidth}
			\centering \small Result with SDC
		\end{subfigure}\\%
		\vspace{1.5mm}%
		\begin{subfigure}[c]{0.45\textwidth}
			\centering
			\includegraphics[width=1\textwidth]{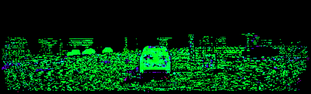}
		\end{subfigure}
		\begin{subfigure}[c]{0.45\textwidth}
			\centering
			\includegraphics[width=1\textwidth]{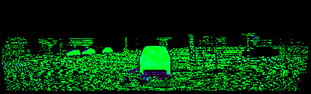}
		\end{subfigure}\\%
		\vspace{0.5mm}%
		\begin{subfigure}[c]{0.45\textwidth}
			\centering \small Error Map for Original Result
		\end{subfigure}
		\begin{subfigure}[c]{0.45\textwidth}
			\centering \small Error Map for Result with SDC
		\end{subfigure}		
		\caption{Optical flow result for CPM \cite{hu2016efficient} on KITTI \cite{menze2015object}. Quantitative results are given in Table 4 of the main paper.}
		\label{fig:visual_results:cpm_kitti}
	\end{subfigure}\\%
	\vspace{10mm}%
	\begin{subfigure}[c]{1.0\textwidth}
		\centering
		\begin{subfigure}[c]{0.45\textwidth}
			\centering
			\includegraphics[width=1\textwidth]{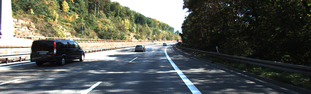}
		\end{subfigure}
		\begin{subfigure}[c]{0.45\textwidth}
			\centering
			\includegraphics[width=1\textwidth]{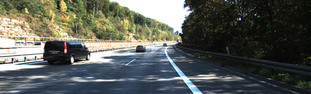}
		\end{subfigure}\\%
		\vspace{0.5mm}%
		\begin{subfigure}[c]{0.45\textwidth}
			\centering \small First Image
		\end{subfigure}
		\begin{subfigure}[c]{0.45\textwidth}
			\centering \small Second Image
		\end{subfigure}\\%
		\vspace{1.5mm}%
		\begin{subfigure}[c]{0.45\textwidth}
			\centering
			\includegraphics[width=1\textwidth]{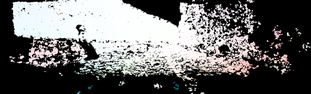}
		\end{subfigure}
		\begin{subfigure}[c]{0.45\textwidth}
			\centering
			\includegraphics[width=1\textwidth]{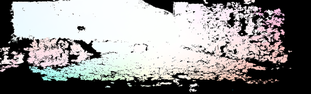}
		\end{subfigure}\\%
		\vspace{0.5mm}%
		\begin{subfigure}[c]{0.45\textwidth}
			\centering \small Original Result
		\end{subfigure}
		\begin{subfigure}[c]{0.45\textwidth}
			\centering \small Result with SDC
		\end{subfigure}\\%
		\vspace{1.5mm}%
		\begin{subfigure}[c]{0.45\textwidth}
			\centering
			\includegraphics[width=1\textwidth]{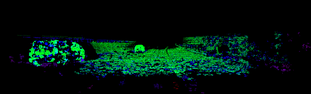}
		\end{subfigure}
		\begin{subfigure}[c]{0.45\textwidth}
			\centering
			\includegraphics[width=1\textwidth]{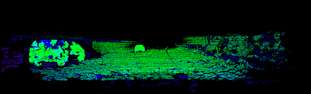}
		\end{subfigure}\\%
		\vspace{0.5mm}%
		\begin{subfigure}[c]{0.45\textwidth}
			\centering \small Error Map for Original Result
		\end{subfigure}
		\begin{subfigure}[c]{0.45\textwidth}
			\centering \small Error Map for Result with SDC
		\end{subfigure}		
		\caption{Optical flow result for FlowFields++ \cite{schuster2018ffpp} on KITTI \cite{menze2015object}. Quantitative results are given in Table 3 of the main paper.}
		\label{fig:visual_results:ffpp_kitti}
	\end{subfigure}
\end{figure*}

\begin{figure*}[p]
	\ContinuedFloat
	\centering
	\begin{subfigure}[c]{1.0\textwidth}
		\centering
		\begin{subfigure}[c]{0.45\textwidth}
			\centering
			\includegraphics[width=1\textwidth]{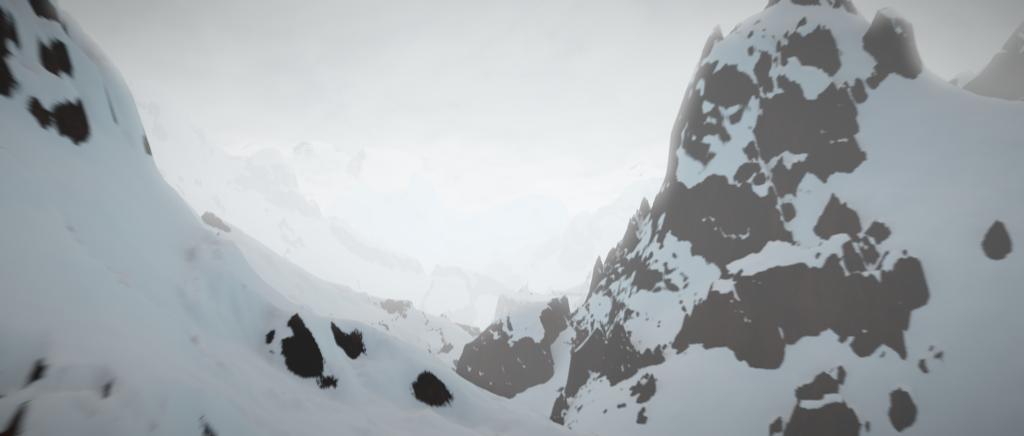}
		\end{subfigure}
		\begin{subfigure}[c]{0.45\textwidth}
			\centering
			\includegraphics[width=1\textwidth]{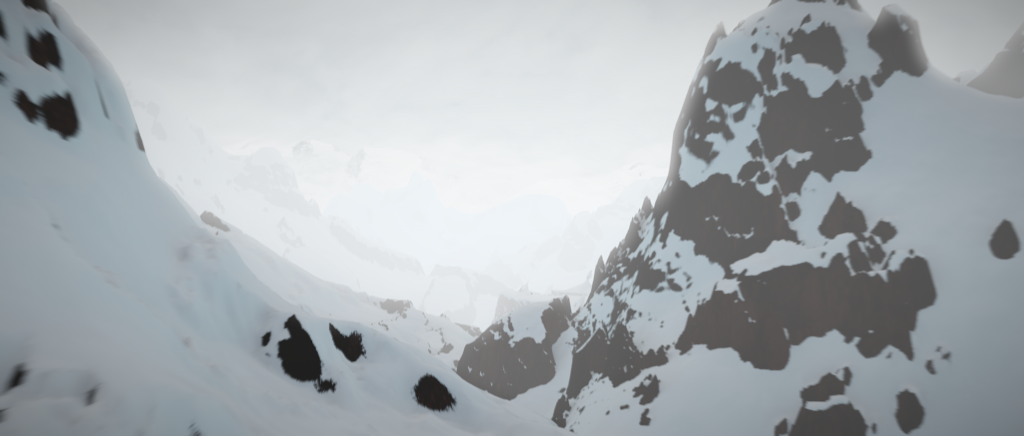}
		\end{subfigure}\\%
		\vspace{0.5mm}%
		\begin{subfigure}[c]{0.45\textwidth}
			\centering \small Left Image
		\end{subfigure}
		\begin{subfigure}[c]{0.45\textwidth}
			\centering \small Right Image
		\end{subfigure}\\%
		\vspace{1.5mm}%
		\begin{subfigure}[c]{0.45\textwidth}
			\centering
			\includegraphics[width=1\textwidth]{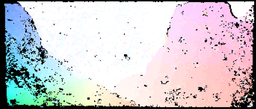}
		\end{subfigure}
		\begin{subfigure}[c]{0.45\textwidth}
			\centering
			\includegraphics[width=1\textwidth]{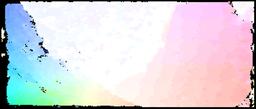}
		\end{subfigure}\\%
		\vspace{0.5mm}%
		\begin{subfigure}[c]{0.45\textwidth}
			\centering \small Original Result
		\end{subfigure}
		\begin{subfigure}[c]{0.45\textwidth}
			\centering \small Result with SDC
		\end{subfigure}\\%
		\vspace{1.5mm}%
		\begin{subfigure}[c]{0.45\textwidth}
			\centering
			\includegraphics[width=1\textwidth]{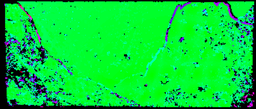}
		\end{subfigure}
		\begin{subfigure}[c]{0.45\textwidth}
			\centering
			\includegraphics[width=1\textwidth]{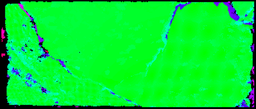}
		\end{subfigure}\\%
		\vspace{0.5mm}%
		\begin{subfigure}[c]{0.45\textwidth}
			\centering \small Error Map for Original Result
		\end{subfigure}
		\begin{subfigure}[c]{0.45\textwidth}
			\centering \small Error Map for Result with SDC
		\end{subfigure}		
		\caption{Optical flow result for CPM \cite{hu2016efficient} on Sintel \cite{butler2012sintel}. Quantitative results are given in Table 4 of the main paper.}
		\label{fig:visual_results:cpm_sintel}
	\end{subfigure}
\end{figure*}

\begin{figure*}[p]
	\ContinuedFloat
	\centering
	\begin{subfigure}[c]{1.0\textwidth}
		\centering
		\begin{subfigure}[c]{0.45\textwidth}
			\centering
			\includegraphics[width=1\textwidth]{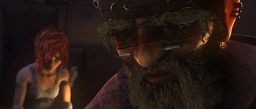}
		\end{subfigure}
		\begin{subfigure}[c]{0.45\textwidth}
			\centering
			\includegraphics[width=1\textwidth]{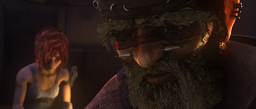}
		\end{subfigure}\\%
		\vspace{0.5mm}%
		\begin{subfigure}[c]{0.45\textwidth}
			\centering \small Left Image
		\end{subfigure}
		\begin{subfigure}[c]{0.45\textwidth}
			\centering \small Right Image
		\end{subfigure}\\%
		\vspace{1.5mm}%
		\begin{subfigure}[c]{0.45\textwidth}
			\centering
			\includegraphics[width=1\textwidth]{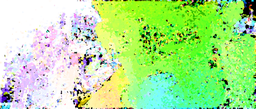}
		\end{subfigure}
		\begin{subfigure}[c]{0.45\textwidth}
			\centering
			\includegraphics[width=1\textwidth]{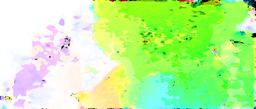}
		\end{subfigure}\\%
		\vspace{0.5mm}%
		\begin{subfigure}[c]{0.45\textwidth}
			\centering \small Original Result
		\end{subfigure}
		\begin{subfigure}[c]{0.45\textwidth}
			\centering \small Result with SDC
		\end{subfigure}\\%
		\vspace{1.5mm}%
		\begin{subfigure}[c]{0.45\textwidth}
			\centering
			\includegraphics[width=1\textwidth]{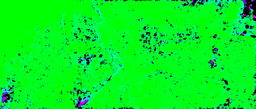}
		\end{subfigure}
		\begin{subfigure}[c]{0.45\textwidth}
			\centering
			\includegraphics[width=1\textwidth]{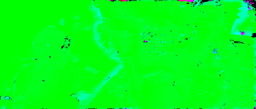}
		\end{subfigure}\\%
		\vspace{0.5mm}%
		\begin{subfigure}[c]{0.45\textwidth}
			\centering \small Error Map for Original Result
		\end{subfigure}
		\begin{subfigure}[c]{0.45\textwidth}
			\centering \small Error Map for Result with SDC
		\end{subfigure}		
		\caption{Optical flow result for FlowFields++ \cite{schuster2018ffpp} on Sintel \cite{butler2012sintel}. Quantitative results are given in Table 3 of the main paper.}
		\label{fig:visual_results:ffpp_sintel}
	\end{subfigure}
\end{figure*}

\begin{figure*}[p]
	\ContinuedFloat
	\centering
	\begin{subfigure}[c]{1.0\textwidth}
		\centering
		\begin{subfigure}[c]{0.45\textwidth}
			\centering
			\includegraphics[width=1\textwidth]{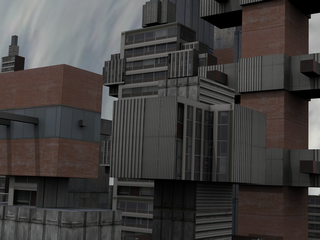}
		\end{subfigure}
		\begin{subfigure}[c]{0.45\textwidth}
			\centering
			\includegraphics[width=1\textwidth]{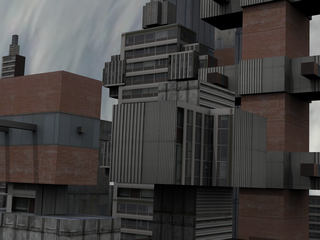}
		\end{subfigure}\\%
		\vspace{0.5mm}%
		\begin{subfigure}[c]{0.45\textwidth}
			\centering \small Left Image
		\end{subfigure}
		\begin{subfigure}[c]{0.45\textwidth}
			\centering \small Right Image
		\end{subfigure}\\%
		\vspace{1.5mm}%
		\begin{subfigure}[c]{0.45\textwidth}
			\centering
			\includegraphics[width=1\textwidth]{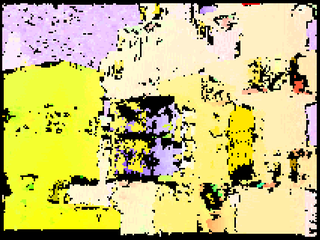}
		\end{subfigure}
		\begin{subfigure}[c]{0.45\textwidth}
			\centering
			\includegraphics[width=1\textwidth]{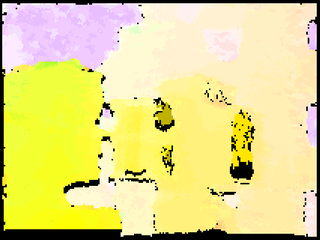}
		\end{subfigure}\\%
		\vspace{0.5mm}%
		\begin{subfigure}[c]{0.45\textwidth}
			\centering \small Original Result
		\end{subfigure}
		\begin{subfigure}[c]{0.45\textwidth}
			\centering \small Result with SDC
		\end{subfigure}\\%
		\vspace{1.5mm}%
		\begin{subfigure}[c]{0.45\textwidth}
			\centering
			\includegraphics[width=1\textwidth]{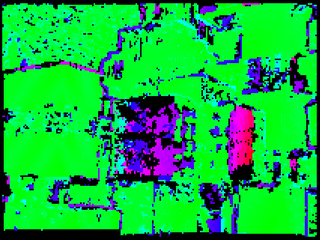}
		\end{subfigure}
		\begin{subfigure}[c]{0.45\textwidth}
			\centering
			\includegraphics[width=1\textwidth]{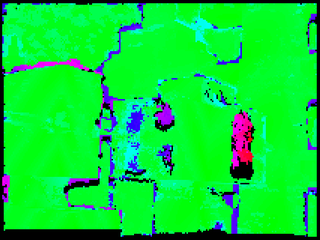}
		\end{subfigure}\\%
		\vspace{0.5mm}%
		\begin{subfigure}[c]{0.45\textwidth}
			\centering \small Error Map for Original Result
		\end{subfigure}
		\begin{subfigure}[c]{0.45\textwidth}
			\centering \small Error Map for Result with SDC
		\end{subfigure}		
		\caption{Optical flow result for CPM \cite{hu2016efficient} on Middlebury \cite{baker2011database}. Quantitative results are given in Table 4 of the main paper.}
		\label{fig:visual_results:cpm_middlebury}
	\end{subfigure}
\end{figure*}

\begin{figure*}[p]
	\ContinuedFloat
	\centering
	\begin{subfigure}[c]{1.0\textwidth}
		\centering
		\begin{subfigure}[c]{0.45\textwidth}
			\centering
			\includegraphics[width=1\textwidth]{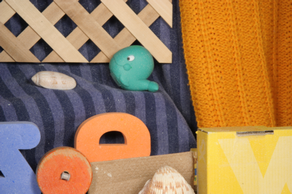}
		\end{subfigure}
		\begin{subfigure}[c]{0.45\textwidth}
			\centering
			\includegraphics[width=1\textwidth]{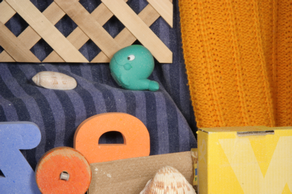}
		\end{subfigure}\\%
		\vspace{0.5mm}%
		\begin{subfigure}[c]{0.45\textwidth}
			\centering \small Left Image
		\end{subfigure}
		\begin{subfigure}[c]{0.45\textwidth}
			\centering \small Right Image
		\end{subfigure}\\%
		\vspace{1.5mm}%
		\begin{subfigure}[c]{0.45\textwidth}
			\centering
			\includegraphics[width=1\textwidth]{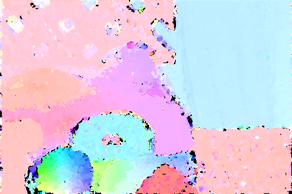}
		\end{subfigure}
		\begin{subfigure}[c]{0.45\textwidth}
			\centering
			\includegraphics[width=1\textwidth]{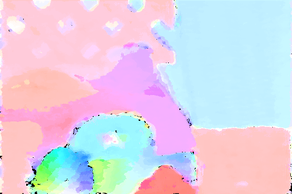}
		\end{subfigure}\\%
		\vspace{0.5mm}%
		\begin{subfigure}[c]{0.45\textwidth}
			\centering \small Original Result
		\end{subfigure}
		\begin{subfigure}[c]{0.45\textwidth}
			\centering \small Result with SDC
		\end{subfigure}\\%
		\vspace{1.5mm}%
		\begin{subfigure}[c]{0.45\textwidth}
			\centering
			\includegraphics[width=1\textwidth]{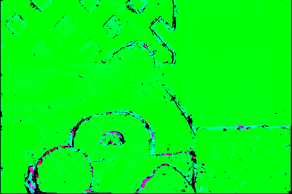}
		\end{subfigure}
		\begin{subfigure}[c]{0.45\textwidth}
			\centering
			\includegraphics[width=1\textwidth]{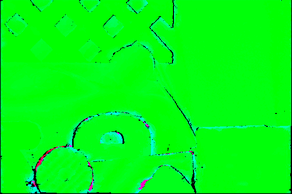}
		\end{subfigure}\\%
		\vspace{0.5mm}%
		\begin{subfigure}[c]{0.45\textwidth}
			\centering \small Error Map for Original Result
		\end{subfigure}
		\begin{subfigure}[c]{0.45\textwidth}
			\centering \small Error Map for Result with SDC
		\end{subfigure}		
		\caption{Optical flow result for FlowFields++ \cite{schuster2018ffpp} on Middlebury \cite{baker2011database}. Quantitative results are given in Table 3 of the main paper.}
		\label{fig:visual_results:ffpp_middlebury}
	\end{subfigure}
\end{figure*}

\begin{figure*}[p]
	\ContinuedFloat
	\centering
	\begin{subfigure}[c]{1.0\textwidth}
		\centering
		\begin{subfigure}[c]{0.45\textwidth}
			\centering
			\includegraphics[width=1\textwidth]{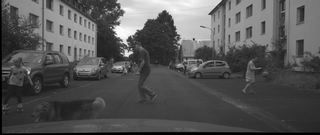}
		\end{subfigure}
		\begin{subfigure}[c]{0.45\textwidth}
			\centering
			\includegraphics[width=1\textwidth]{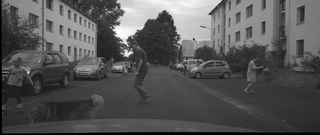}
		\end{subfigure}\\%
		\vspace{0.5mm}%
		\begin{subfigure}[c]{0.45\textwidth}
			\centering \small Left Image
		\end{subfigure}
		\begin{subfigure}[c]{0.45\textwidth}
			\centering \small Right Image
		\end{subfigure}\\%
		\vspace{1.5mm}%
		\begin{subfigure}[c]{0.45\textwidth}
			\centering
			\includegraphics[width=1\textwidth]{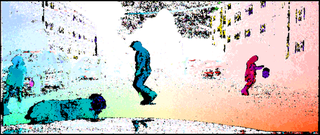}
		\end{subfigure}
		\begin{subfigure}[c]{0.45\textwidth}
			\centering
			\includegraphics[width=1\textwidth]{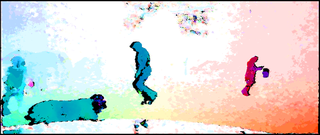}
		\end{subfigure}\\%
		\vspace{0.5mm}%
		\begin{subfigure}[c]{0.45\textwidth}
			\centering \small Original Result
		\end{subfigure}
		\begin{subfigure}[c]{0.45\textwidth}
			\centering \small Result with SDC
		\end{subfigure}\\%
		\vspace{1.5mm}%
		\begin{subfigure}[c]{0.45\textwidth}
			\centering
			\includegraphics[width=1\textwidth]{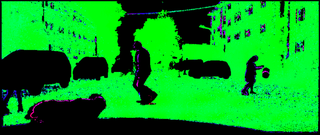}
		\end{subfigure}
		\begin{subfigure}[c]{0.45\textwidth}
			\centering
			\includegraphics[width=1\textwidth]{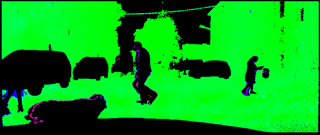}
		\end{subfigure}\\%
		\vspace{0.5mm}%
		\begin{subfigure}[c]{0.45\textwidth}
			\centering \small Error Map for Original Result
		\end{subfigure}
		\begin{subfigure}[c]{0.45\textwidth}
			\centering \small Error Map for Result with SDC
		\end{subfigure}		
		\caption{Optical flow result CPM \cite{hu2016efficient} on HD1K \cite{kondermann2016hci}. Quantitative results are given in Table 4 of the main paper.}
		\label{fig:visual_results:cpm_HD1K}
	\end{subfigure}
\end{figure*}

\begin{figure*}[p]
	\ContinuedFloat
	\centering
	\begin{subfigure}[c]{1.0\textwidth}
		\centering
		\begin{subfigure}[c]{0.45\textwidth}
			\centering
			\includegraphics[width=1\textwidth]{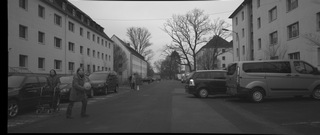}
		\end{subfigure}
		\begin{subfigure}[c]{0.45\textwidth}
			\centering
			\includegraphics[width=1\textwidth]{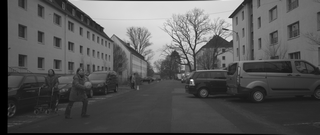}
		\end{subfigure}\\%
		\vspace{0.5mm}%
		\begin{subfigure}[c]{0.45\textwidth}
			\centering \small Left Image
		\end{subfigure}
		\begin{subfigure}[c]{0.45\textwidth}
			\centering \small Right Image
		\end{subfigure}\\%
		\vspace{1.5mm}%
		\begin{subfigure}[c]{0.45\textwidth}
			\centering
			\includegraphics[width=1\textwidth]{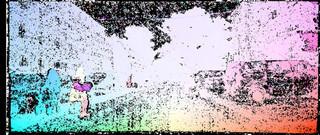}
		\end{subfigure}
		\begin{subfigure}[c]{0.45\textwidth}
			\centering
			\includegraphics[width=1\textwidth]{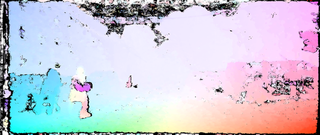}
		\end{subfigure}\\%
		\vspace{0.5mm}%
		\begin{subfigure}[c]{0.45\textwidth}
			\centering \small Original Result
		\end{subfigure}
		\begin{subfigure}[c]{0.45\textwidth}
			\centering \small Result with SDC
		\end{subfigure}\\%
		\vspace{1.5mm}%
		\begin{subfigure}[c]{0.45\textwidth}
			\centering
			\includegraphics[width=1\textwidth]{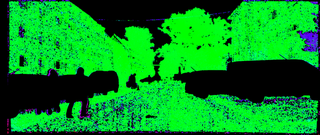}
		\end{subfigure}
		\begin{subfigure}[c]{0.45\textwidth}
			\centering
			\includegraphics[width=1\textwidth]{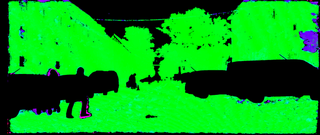}
		\end{subfigure}\\%
		\vspace{0.5mm}%
		\begin{subfigure}[c]{0.45\textwidth}
			\centering \small Error Map for Original Result
		\end{subfigure}
		\begin{subfigure}[c]{0.45\textwidth}
			\centering \small Error Map for Result with SDC
		\end{subfigure}		
		\caption{Optical flow result for FlowFields++ \cite{schuster2018ffpp} on HD1K \cite{kondermann2016hci}. Quantitative results are given in Table 3 of the main paper.}
		\label{fig:visual_results:ffpp_hd1k}
	\end{subfigure}
\end{figure*}

\begin{figure*}[p]
	\ContinuedFloat
	\centering
	\begin{subfigure}[c]{1.0\textwidth}
		\centering
		\begin{subfigure}[c]{0.45\textwidth}
			\centering
			\includegraphics[width=1\textwidth]{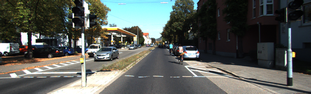}
		\end{subfigure}
		\begin{subfigure}[c]{0.45\textwidth}
			\centering
			\includegraphics[width=1\textwidth]{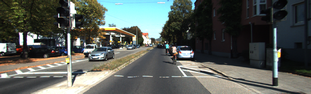}
		\end{subfigure}\\%
		\vspace{0.5mm}%
		\begin{subfigure}[c]{0.45\textwidth}
			\centering \small First Left Image
		\end{subfigure}
		\begin{subfigure}[c]{0.45\textwidth}
			\centering \small First Right Image
		\end{subfigure}\\%
		\vspace{1.5mm}%
		\begin{subfigure}[c]{0.45\textwidth}
			\centering
			\includegraphics[width=1\textwidth]{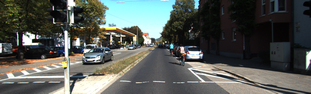}
		\end{subfigure}
		\begin{subfigure}[c]{0.45\textwidth}
			\centering
			\includegraphics[width=1\textwidth]{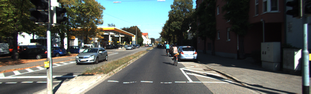}
		\end{subfigure}\\%
		\vspace{0.5mm}%
		\begin{subfigure}[c]{0.45\textwidth}
			\centering \small Second Left Image
		\end{subfigure}
		\begin{subfigure}[c]{0.45\textwidth}
			\centering \small Second Right Image
		\end{subfigure}\\%
		\vspace{1.5mm}%
		\begin{subfigure}[c]{0.45\textwidth}
			\centering
			\includegraphics[width=1\textwidth]{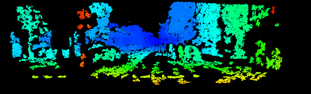}
		\end{subfigure}
		\begin{subfigure}[c]{0.45\textwidth}
			\centering
			\includegraphics[width=1\textwidth]{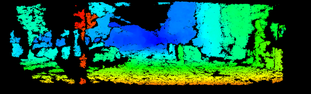}
		\end{subfigure}\\%
		\vspace{0.5mm}%
		\begin{subfigure}[c]{0.45\textwidth}
			\centering \small Original Result, Disparity 1
		\end{subfigure}
		\begin{subfigure}[c]{0.45\textwidth}
			\centering \small Result with SDC, Disparity 1
		\end{subfigure}\\%
		\vspace{1.5mm}%
		\begin{subfigure}[c]{0.45\textwidth}
			\centering
			\includegraphics[width=1\textwidth]{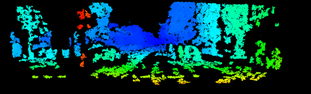}
		\end{subfigure}
		\begin{subfigure}[c]{0.45\textwidth}
			\centering
			\includegraphics[width=1\textwidth]{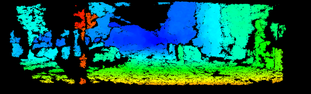}
		\end{subfigure}\\%
		\vspace{0.5mm}%
		\begin{subfigure}[c]{0.45\textwidth}
			\centering \small Original Result, Disparity 2
		\end{subfigure}
		\begin{subfigure}[c]{0.45\textwidth}
			\centering \small Result with SDC, Disparity 2
		\end{subfigure}\\%
		\vspace{1.5mm}%
		\begin{subfigure}[c]{0.45\textwidth}
			\centering
			\includegraphics[width=1\textwidth]{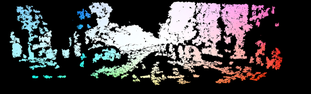}
		\end{subfigure}
		\begin{subfigure}[c]{0.45\textwidth}
			\centering
			\includegraphics[width=1\textwidth]{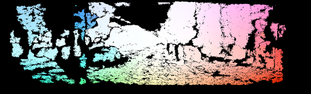}
		\end{subfigure}\\%
		\vspace{0.5mm}%
		\begin{subfigure}[c]{0.45\textwidth}
			\centering \small Original Result, Optical Flow
		\end{subfigure}
		\begin{subfigure}[c]{0.45\textwidth}
			\centering \small Result with SDC, Optical Flow
		\end{subfigure}\\%
		\vspace{1.5mm}%
		\begin{subfigure}[c]{0.45\textwidth}
			\centering
			\includegraphics[width=1\textwidth]{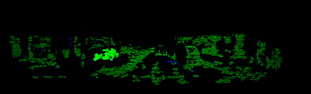}
		\end{subfigure}
		\begin{subfigure}[c]{0.45\textwidth}
			\centering
			\includegraphics[width=1\textwidth]{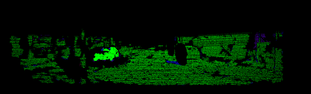}
		\end{subfigure}\\%
		\vspace{0.5mm}%
		\begin{subfigure}[c]{0.45\textwidth}
			\centering \small Error Map for Original Result
		\end{subfigure}
		\begin{subfigure}[c]{0.45\textwidth}
			\centering \small Error Map for Result with SDC
		\end{subfigure}		
		\caption{Scene flow result for SceneFlowFields \cite{schuster2018sceneflowfields} on KITTI \cite{menze2015object}. Quantitative results are given in Table 5 of the main paper.}
		\label{fig:visual_results:sff_kitti}
	\end{subfigure}
	\caption{Visual comparison of different matching algorithms for stereo disparity, optical flow, and scene flow on different data sets. For each combination, we show the original matching result and the results using our SDC feature descriptor. We do not change anything but the feature descriptor. With our SDC feature network, matching is more accurate (less outliers, sharper boundaries, smoother surface areas) and much denser (more matches over the complete image).}
	\label{fig:visual_results}
\end{figure*}

{\small
\bibliographystyle{ieee_fullname}
\bibliography{bib}
}